\documentclass[10pt,twocolumn,letterpaper]{article}

\usepackage[pagenumbers]{iccv} 
\usepackage{times}
\usepackage{epsfig}
\usepackage{graphicx}
\usepackage{amsmath}
\usepackage{amssymb}
\usepackage{subcaption}
\usepackage{wrapfig} 
\usepackage{float}
\usepackage{pifont}
\usepackage{booktabs}
\usepackage{multirow}
\usepackage{xcolor}
\newcommand{\E}{\mathbb{E}}
\definecolor{iccvblue}{rgb}{0.21,0.49,0.74}
\usepackage[pagebackref,breaklinks,colorlinks,allcolors=iccvblue]{hyperref}

\def\paperID{6101} 
\def\confName{ICCV}
\def\confYear{2025}

\title{GT-Mean Loss: A Simple Yet Effective Solution for Brightness Mismatch in Low-Light Image Enhancement}

\author{Jingxi Liao, Shijie Hao\thanks{Corresponding author}, Richang Hong, Meng Wang, \\
Hefei University of Technology, Hefei, China\\
{\tt\small \{jingxi.ljx, hfut.hsj, hongrc.hfut, eric.mengwang\}@gmail.com}
}
\newif\ifappendix
\appendixfalse  
\appendixtrue  
\begin{document}
\maketitle

\begin{abstract}
Low-light image enhancement (LLIE) aims to improve the visual quality of images captured under poor lighting conditions. In supervised LLIE research, there exists a significant yet often overlooked inconsistency between the overall brightness of an enhanced image and its ground truth counterpart, referred to as \textit{brightness mismatch} in this study. Brightness mismatch negatively impact supervised LLIE models by misleading model training. However, this issue is largely neglected in current research. 
In this context, we propose the \textit{GT-mean loss}, a simple yet effective loss function directly modeling the mean values of images from a probabilistic perspective.
The GT-mean loss is flexible, as it extends existing supervised LLIE loss functions into the GT-mean form with minimal additional computational costs. Extensive experiments demonstrate that the incorporation of the GT-mean loss results in consistent performance improvements across various methods and datasets. 
\renewcommand{\thefootnote}{}
\footnotetext{This work was supported by the National Natural Science Foundation of China under Grant 62172137.}

\end{abstract}


\section{INTRODUCTION}

Low-light image enhancement (LLIE) is a crucial task in computer vision, aiming to improve the overall quality of images captured under poor lighting conditions ~\cite{9609683,liu2021bench}. The primary objective of training a supervised LLIE model, denoted as $f(\cdot)$, is to map a low-light image $x$ to an enhanced image $f(x)$, subjecting to the constraint that $f(x)$ should resemble the ground truth (GT) image $y$ as much as possible. While existing methods achieve reasonable brightness adjustment, they face a critical dilemma: models disproportionately prioritize global brightness alignment at the expense of suppressing other degradation factors like noise ~\cite{9730802, Wei_2020_CVPR, Moseley_2021_CVPR}, color distortion ~\cite{feng2024hvi, Zhang_2022_CVPR}, and artifacts~\cite{zhou2022lednet, Zhou_2021_ICCV}. This misalignment stems from three interrelated issues:  

\begin{itemize}
\item \textbf{Inevitable Brightness Residual}: The inconsistency in overall brightness between the supervised low-light image enhancement (LLIE) function $f(x)$ and the target image $y$ is a common occurrence. The inherent nonlinear mapping from low-light to well-lit images inevitably leaves behind a small brightness residual, which can be simply expressed as $\mathbb{E}[f(x)] \neq \mathbb{E}[y]$. We refer to this phenomenon as brightness mismatch.

\item \textbf{Sensitivity of Loss Functions}: Pixel-level losses (e.g., L1/L2) lack a perceptual saturation mechanism—they penalize brightness discrepancy regardless of whether these mismatches are visually negligible once  $\mathbb{E}[f(x)] \neq \mathbb{E}[y]$.

\item \textbf{Misalignment with Human Perception}: Under adequate brightness, human vision prioritizes noise suppression and color fidelity \cite{Brown_49,weber}, yet these factors contribute far less to pixel-level losses than brightness residuals.
\end{itemize}

\begin{figure}[h]  
\centering
\includegraphics[width=0.48\textwidth]{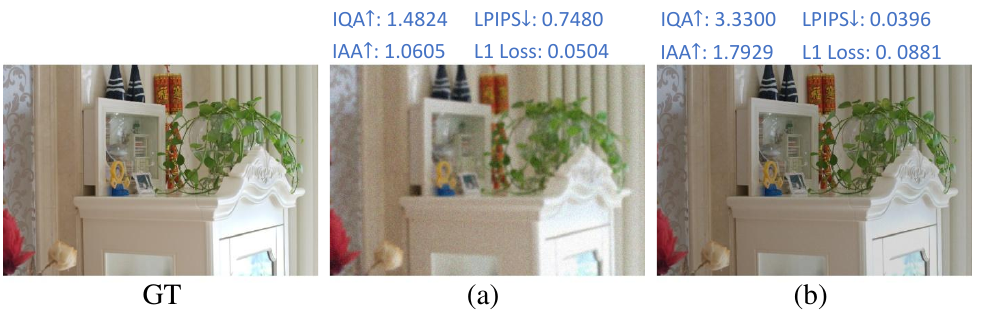} 
\caption{(a) exhibits severe degradation (noise and blur), with overall brightness consistent with the GT, while (b) has mostly clear details and textures, with overall brightness at 0.9 times that of the GT. We use LPIPS($\downarrow$)~\cite{8578166} , QALIGN's~\cite{wu24ah} IQA($\uparrow$) and IAA($\uparrow$) modules to represent the image quality of (a) and (b). This example reveals an unexpected behavior of the L1 loss (a typical pixel-level loss): the L1 loss assigns a lower loss value to the low-quality image (a) and a higher loss value to the high-quality image (b). }  
\label{fig2}
\end{figure}

As a result, the model disproportionately pursues marginal benefits in brightness alignment, entering a futile optimization cycle. This misguided effort directly contradicts the priorities of human vision—once basic brightness is met, noise and color accuracy dominate perceptual quality. Figure~\ref{fig2} illustrates a typical contradictory scenario.
\begin{figure*}[h]
	\centering
	\begin{subfigure}[b]{0.25\textwidth}
		\centering
		\includegraphics[width=\textwidth]{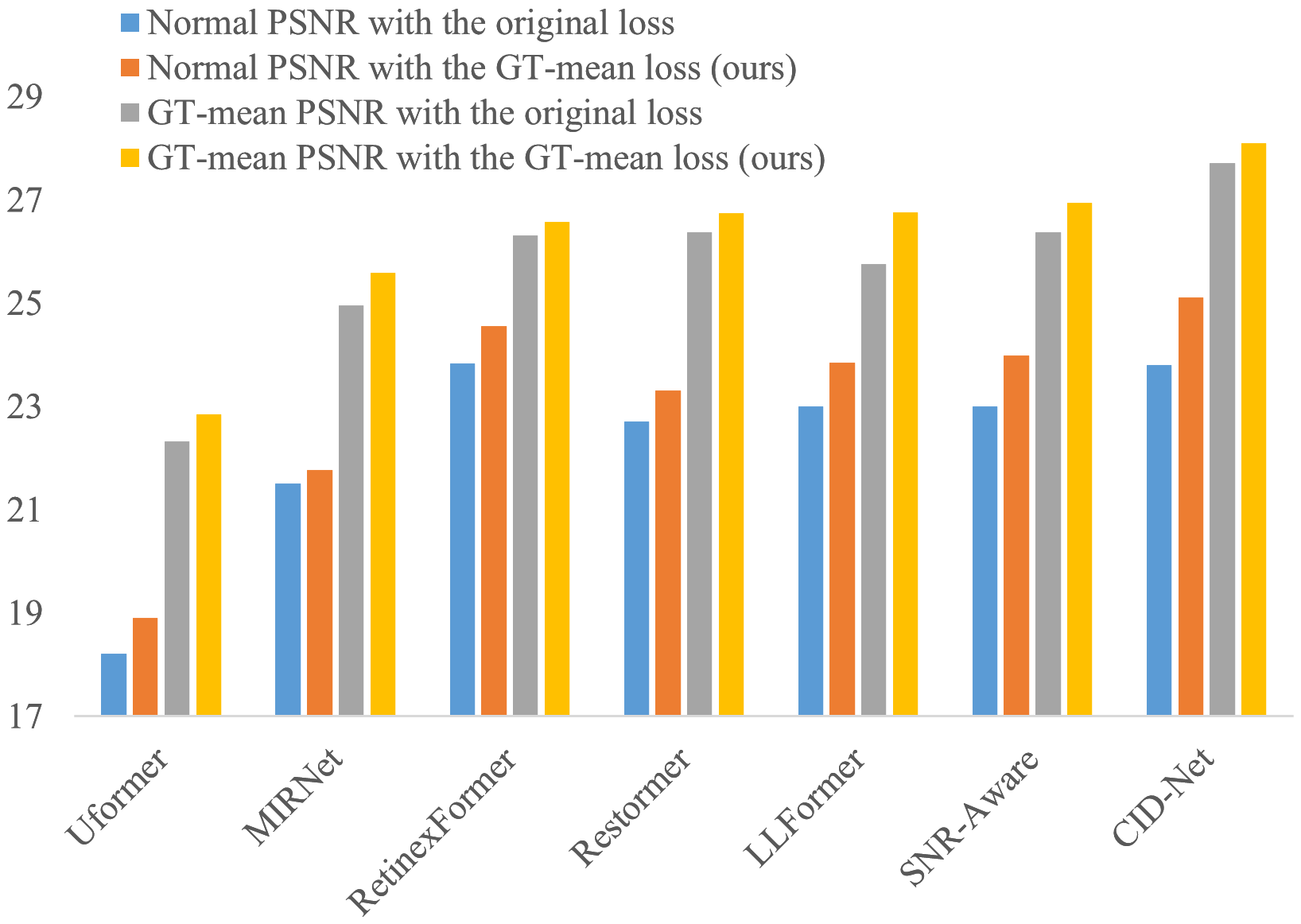}
		\caption{LOLv1}
	\end{subfigure}
	\begin{subfigure}[b]{0.25\textwidth}
		\centering
		\includegraphics[width=\textwidth]{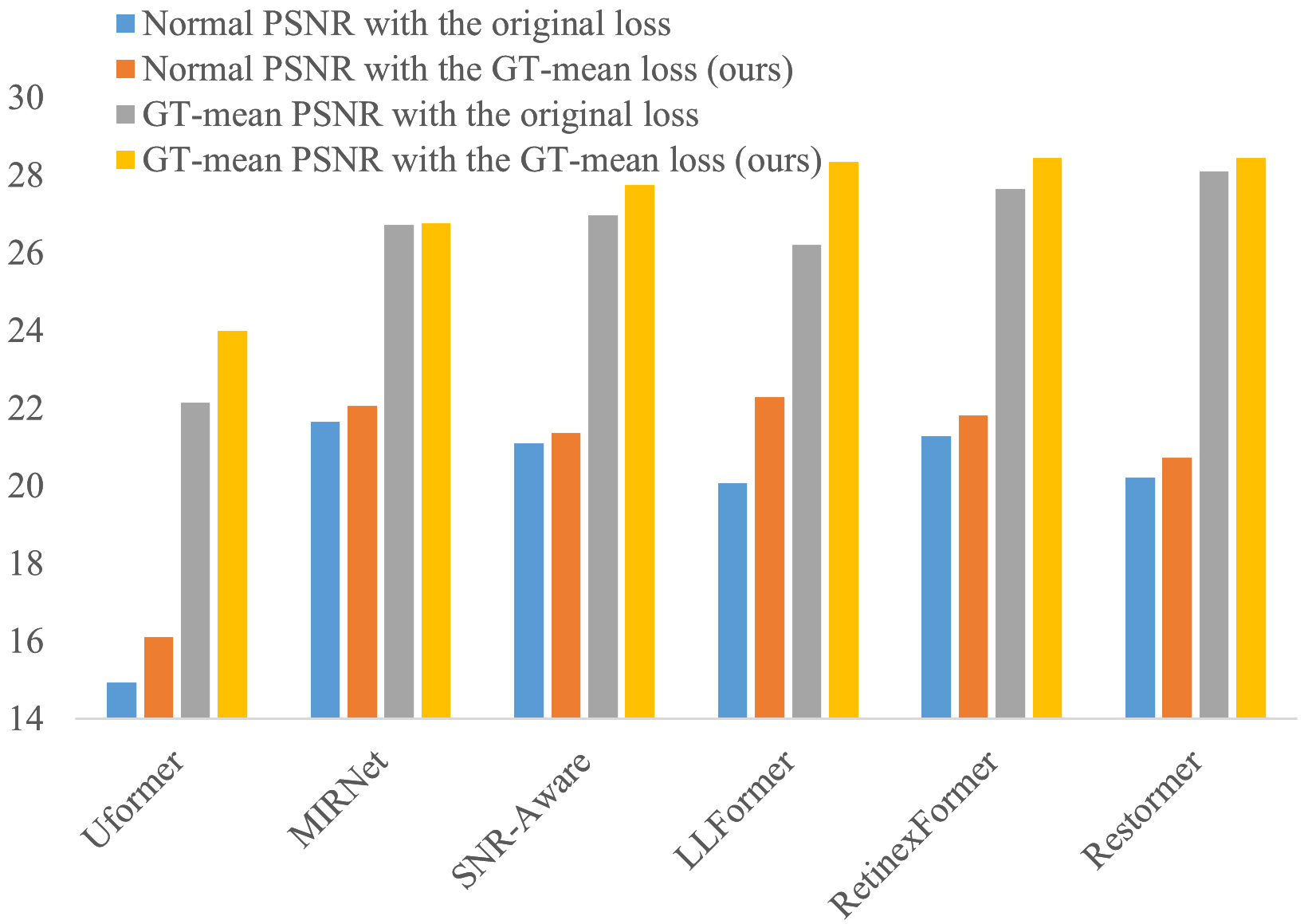}
		\caption{LOLv2 real}
	\end{subfigure}
	\begin{subfigure}[b]{0.25\textwidth}
		\centering
		\includegraphics[width=\textwidth]{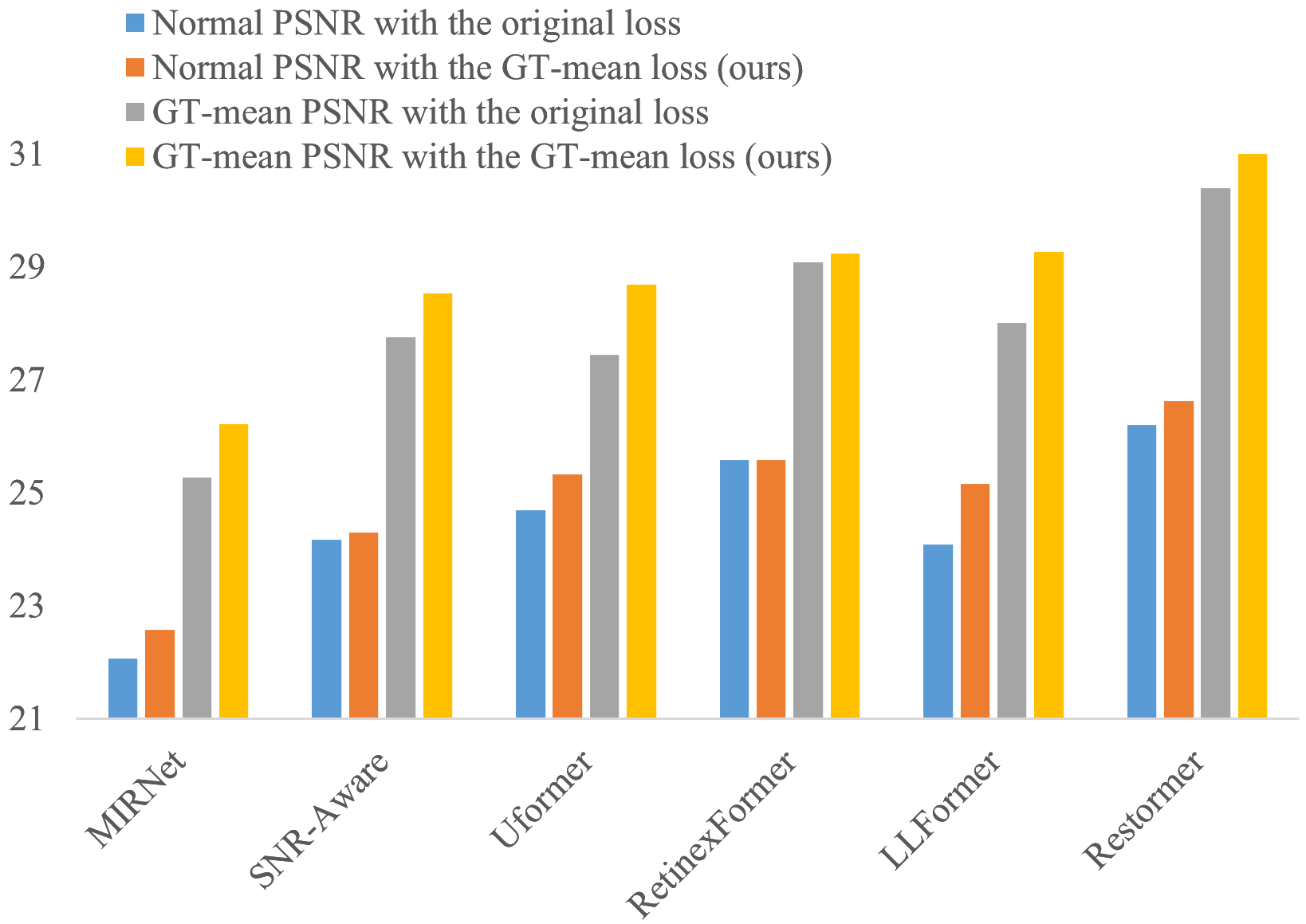}
		\caption{LOLv2 synthetic }
	\end{subfigure}
	\caption{Performance across various supervised LLIE models trained by their original loss functions and our GT-mean loss functions. The performance is consistently improved when the GT-mean loss functions are adopted. Notably, this improvement is easily attainable, as the use of GT-mean loss functions is flexible and brings minimal additional computational costs during training.}
	\label{fig1}
\end{figure*}\\
The issue under brightness mismatch has largely been ignored in existing supervised LLIE research, despite some indirect solutions that do not primarily address this problem. For example, ~\cite{Chen_Wang_Yang_Liu_2018, Yang_Wang_Huang_Wang_Liu_2021, Wu_2022_CVPR,9844872} designed multiple sub-networks to decouple brightness from other factors and optimized them separately. Nevertheless, the divide-and-conquer roadmap inevitably complicates the model design, as well as introducing significant computational overhead. Others use perceptual losses~\cite{Johnson_Alahi_FeiFei_2016} to bypass pixel-level alignment, sacrificing reconstruction fidelity. A unified, efficient solution remains elusive.

Inspired by human visual adaptation, we propose the GT-Mean Loss, which dynamically balances two optimization objectives to enable end-to-end training:

\begin{itemize}
\item \textbf{Original Loss Term}.  
Maintain the model’s baseline capabilities by directly optimizing \( L(f(x), y) \).

\item \textbf{Brightness-Aligned Loss Term}.  
This term is defined as \( L(\lambda f(x), y) \), where \( \lambda = \mathbb{E}[y] / \mathbb{E}[f(x)] \) linearly scales \( f(x) \) such that the mean brightness of the scaled output \( \mathbb{E}[\lambda f(x)] \) aligns with \( \mathbb{E}[ y] \) . This eliminates the interference of brightness differences during optimization.
\end{itemize}

To dynamically balance these two terms, we introduce \( W \), which represents the Bhattacharyya distance between the brightness distributions of \( f(x) \) and \( y \). Unlike multi-stage frameworks, GT-Mean Loss enables brightness-decoupled optimization without requiring architectural modifications or manually tuned thresholds.

Although perceptual losses can to some extent ignore the impact of brightness mismatch, almost all perceptual losses abandon corrections. In contrast, our GT-mean loss directly addresses the influence of brightness mismatch on pixel-level loss.  The results in our experimental section validate this point in Section~\ref{4_2}.

In summary, the GT-mean loss is highlighted in the following aspects:

\begin{itemize}
	\item \textbf{Simplicity}: The construction of the GT-mean loss is both theoretically and practically straightforward. Its underlying mechanism is easy to understand, and its implementation is uncomplicated.
	
	\item \textbf{Flexibility}: The GT-mean loss is highly flexible. For instance, the $L_1$ loss can be directly extended into the $L_1$ GT-mean version. This character makes adopting the GT-mean loss a universal choice for supervised LLIE models to upgrade their loss functions.
	
	\item \textbf{Low Cost}: Using the GT-mean loss introduces minimal overhead during training (approximately doubling the original loss computation). It is negligible compared to the overall model optimization process.
	
	\item \textbf{Effectiveness}: Extensive experiments have demonstrated that the GT-mean loss consistently improves model performance across a wide range of supervised LLIE methods (as shown in Figure \ref{fig1}).
	
\end{itemize}

\section{Related works}
Loss function is essential in LLIE tasks, as it directs model training.  We categorize the loss functions in LLIE into two groups based on their purposes.\\
\textbf{Fidelity Losses.} These losses are designed to ensure that the enhanced image $f(x)$ closely resembles the ground truth $y$. For example, the $L_1$-like loss functions directly ensure the pixel-level fidelity ~\cite{9609683,liu2021bench}. Loss functions focused on color representation often use color histogram-based metrics ~\cite{feng2024hvi}, while others preserve fidelity at the frequency domain ~\cite{wang2023fourllie,huang2022deep}. Among these, pixel-level loss functions are essential for image reconstruction, but their inherent point-to-point computation can easily misinterpret slight brightness variations as image degradations. Additionally, some loss functions aim to maintain fidelity at higher representation levels, such as the perceptual loss ~\cite{Johnson_Alahi_FeiFei_2016}. Recently, some novel loss functions have emerged that subtly utilize fine-grained semantic information ~\cite{Liang_Li_Zhou_Feng_Loy_2023, Wu_Pan_Wang_Yang_Wei_Li_Shen_2023}. While these high-level losses demonstrate effectiveness in perceptual enhancement, they cannot fully substitute pixel-level. Current implementations must be combined with pixel-based losses to ensure functional model performance .\\
While these high-level losses demonstrate effectiveness in semantic-aware enhancement, they cannot fully substitute pixel-level supervision. \\
\textbf{Prior-Based Losses.} These losses integrate domain-specific prior knowledge into LLIE models, aiming to maximize the use of available information. Typical examples are the Retinex-based methods~\cite{b22YonghuaZhang2019KindlingTD,Chen_Wang_Yang_Liu_2018,Yang_Wang_Huang_Wang_Liu_2021,Wu_2022_CVPR,9844872,Fu_2023_CVPR}, which are founded on the Retinex theory that an image is composed of illumination map and reflectance map. Specific loss functions are employed to penalize the locally smooth properties of illumination map and the lightness-insensitive properties of reflectance map.  While providing clear physical interpretability, the ill-posed nature of Retinex decomposition, for instance, can propagate estimation errors in illumination and reflectance maps. This inherent ambiguity may lead to suboptimal results in practical applications. Furthermore, for unsupervised LLIE models that lack GT images for training, prior-based loss functions are essential for guiding model optimization~\cite{Li_Guo_Chen_2021,ma2022toward,Jiang_2024_ECCV,quadprior}. However, these loss functions can be less flexibility and robust.  For example, ~\cite{ma2022toward,Jiang_2024_ECCV,quadprior} heavily rely on specific network designs. ZeroDCE ~\cite{Li_Guo_Chen_2021} builds an exposure control loss function with a hard threshold, which may result in over-exposure.

To pursue comprehensive visual quality enhancement, LLIE models tend to incorporate multiple loss functions. While this strategy can lead to improved results, it also increases the burden of model design and hyperparameter tuning. More importantly, as the existing loss functions overlook the brightness mismatch factor, the fundamental challenges brought by this factor remain unaddressed.

	

\section{METHOD}
In Section \ref{section_3_1} present the formulation of the GT-mean loss. In Section \ref{section_3_2} detail the crucial component $W$ of this loss function. Section \ref{section_3_3} summarizes the features of the proposed loss.

\subsection{GT-mean loss}

\label{section_3_1}

In general, the GT-mean loss $L_{GT}(f(x), y)$ can be regarded as an extension of the existing loss $L(f(x), y)$ used for LLIE. To deal with the issues arising from brightness mismatch, the key to constructing $L_{GT}(f(x), y)$ is matching the average brightness of $f(x)$ and $y$. Furthermore, $L_{GT}(f(x), y)$ is designed to retain the form and effectiveness of the original loss $L(f(x), y)$. Based on this,  $L_{GT}(f(x), y)$ is formulated as follows:

\begin{align}
	\label{eq1}
	&L_{GT}(f(x), y) = \nonumber\\&W\cdot\underbrace{L(f(x), y)}_{original\ loss}+\ \   (1-W) \cdot \underbrace{L\left(\frac{\E[y]}{\E[f(x)]} f(x), y\right)}_{brightness-aligned\ loss},
\end{align}
where  $\frac{\E[y]}{\E[f(x)]}$ is a scaling factor for aligning the average brightness of $f(x)$ and $y$. To clarify, $\mathbb{E}[y]$ and $\mathbb{E}[f(x)]$ denote the average brightness of the ground truth image $y \in \mathbb{R}^{1 \times 3 \times H \times W}$ and the enhanced image $f(x) \in \mathbb{R}^{1 \times 3 \times H \times W}$, respectively. The weight $W \in [0, 1]$ balances the two terms in $L_{GT}$. It is noted that the choice of $L(\cdot)$ is is arbitrary, provided that it accepts $f(x)$ and $y$ as inputs.

\subsection{Weight Design}
\label{section_3_2}
\subsubsection{Probabilistic Modeling on Average Brightness}
According to the Contrast Sensitivity Function~\cite{Robson66, Bühren2018}, human vision is sensitive to local contrasts (e.g., edges, textures, and intensity variations) but less sensitive to minor shifts in average brightness \(\mathbb{E}[\cdot]\). Moreover, research by Mathys ~\cite{Mathys2011, Mathys2014} has shown that uncertainty in human perceptual behavior is often well modeled by Gaussian distributions, making them a suitable choice for representing brightness perception variability. 

This perceptual characteristic motivates our probabilistic modeling of brightness distributions: by treating \(\mathbb{E}[\cdot]\) as a random variable rather than a fixed value, we represent the inherent uncertainty in human perception of brightness. This modeling approach enhances the controllability of the loss function during training, as it allows us to estimate the weight \(W\) using common metrics like the Kullback-Leibler (KL) divergence or Wasserstein distance. These metrics ensure a smooth transition between the two terms in Eq.~\ref{eq1}, avoiding abrupt changes that could destabilize the optimization process. Consequently, the loss function exhibits good continuity with respect to parameter changes and features flat regions around its minima. These flat regions indicate that small perturbations in brightness do not significantly affect the loss value, making the optimization process more robust (as exemplified in Section~\ref{loss_form_chapter}).

Based on these considerations, we treat \(\mathbb{E}[\cdot]\) as an observation from a random variable \(\widetilde{\mathbb{E}}[\cdot]\) that follows a Gaussian distribution. Specifically, for a single ground truth $y \in \mathbb{R}^{1 \times 3 \times H \times W}$, it can be regarded as an observation from \(\widetilde{\mathbb{E}}[y]\), represented as:
\begin{align} \widetilde{\E}[y] = \alpha \E[y], \quad \alpha \sim \mathcal{N}(1, \sigma_{\alpha}^2), 
	\label{eq2}
\end{align} 
where $\alpha$ determines the probability distribution type of $\widetilde{\E}[y]$, and $\sigma_{\alpha}^2$ defines its spread. Therefore, we have average brightness distribution $p(\widetilde{\E}[y]) = \mathcal{N}(\mu_y,\sigma_{y}^2)$, with $\mu_y= \E[y]$ and $\sigma_{y}=\sigma_{\alpha}\E[y]$. 
Similarly, for a single enhanced image $f(x) \in \mathbb{R}^{1 \times 3 \times H \times W}$, its average brightness \( \mathbb{E}[f(x)] \) can also be regarded as an observation from \( \widetilde{\mathbb{E}}[f(x)] \), represented as:
\begin{align}
\widetilde{\mathbb{E}}[f(x)] = \beta \mathbb{E}[f(x)], \quad \beta \sim \mathcal{N}(1, \sigma_{\beta}^2),
\label{eq3}
\end{align}
where \( \beta \) determines the probability distribution type of \( \widetilde{\mathbb{E}}[f(x)] \), and \( \sigma_{\beta}^2 \) defines its spread. Consequently, we have average brightness distribution \( q(\widetilde{\mathbb{E}}[f(x)]) = \mathcal{N}(\mu_{fx}, \sigma_{fx}^2) \), with \( \mu_{fx} = \mathbb{E}[f(x)] \) and \( \sigma_{fx} = \sigma_{\beta} \mathbb{E}[f(x)] \). 

\subsubsection{Estimation of $W$}

To quantify the divergence between the two distributions, we employ the Bhattacharyya distance \cite{1089532}, which offers a closed-form solution for Gaussian distributions, significantly enhancing computational efficiency. For Gaussian distributions \( p(\widetilde{\mathbb{E}}[y]) = \mathcal{N}(\mu_y, \sigma_y^2) \) and \( q(\widetilde{\mathbb{E}}[f(x)]) = \mathcal{N}(\mu_{fx}, \sigma_{fx}^2) \), the closed-form expression for the Bhattacharyya distance\( D_B \)  is given by:

\begin{align}
D_B(p \Vert q) = \frac{1}{4} \frac{(\mu_y - \mu_{fx})^2}{\sigma_y^2 + \sigma_{fx}^2} + \frac{1}{2} \ln\left( \frac{\sigma_y^2 + \sigma_{fx}^2}{2\sigma_y \sigma_{fx}} \right),
\end{align}
where $\mu_y \triangleq \mathbb{E}[y]$ and $\mu_{fx} \triangleq \mathbb{E}[f(x)]$ denote the expected brightness values, while $\sigma_y^2 = (\sigma_{\alpha}\mathbb{E}[y])^2$ and $\sigma_{fx}^2 = (\sigma_{\beta}\mathbb{E}[f(x)])^2$ characterize distribution spreads derived from Eq.~\eqref{eq2} and Eq.~\eqref{eq3}.

To enforce $W$ as a normalized weighting coefficient, we apply value clipping:
\begin{align}
W \leftarrow \text{clip}\left(D_B(p \Vert q), 0, 1\right)
\end{align}
This operation induces three interpretable regimes:
\begin{itemize}
 \item $W=1$: Maximum emphasis on general quality when $\widetilde{\mathbb{E}}[y]$ and $\widetilde{\mathbb{E}}[f(x)]$ exhibit strong divergence ($D_B \geq 1$).
 \item $W=0$: Exclusive focus on brightness-aligned refinement when distributions coincide ($D_B = 0$).
 \item Smooth transition via \( W \in (0,1) \) during intermediate phases .
\end{itemize}

In supervised LLIE, the enhanced image $f(x)$ should closely resemble the ground truth image $y$. Consequently, we assume that the shape of $p(\widetilde{\E}[y])$ and $q(\widetilde{\E}[f(x)])$ are similar. Based on this assumption, we equate $\sigma_{\alpha}^2$ and $\sigma_{\beta}^2$, setting them both to $\sigma^2$. In our experiments, we empirically set $\sigma$ as 0.1 for all the comparisons. In Section \ref{Ablation Study}, we delve deeper into the influence of $\sigma$ on the performance of LLIE.

\subsection{Discussion of the GT-mean loss}
\label{section_3_3}

The primary strength of the GT-mean loss lies in its ability to dynamically balance the model's focus during training. In the early stages, when the difference between \( p(\widetilde{\mathbb{E}}[y]) \) and \( q(\widetilde{\mathbb{E}}[f(x)]) \) is significant, the weight \( W \) is expected to approach 1, causing the first term \( L(f(x), y) \) to dominate the overall loss function. This behavior ensures that the GT-mean loss closely resembles the original loss \( L(f(x), y) \),  thereby preserving the baseline model's capabilities.

As training progresses and \( p(\widetilde{\mathbb{E}}[y]) \) and \( q(\widetilde{\mathbb{E}}[f(x)]) \) become closer, \( W \) decreases to a smaller value. This shift in weight emphasizes the second term \( L\left(\frac{\mathbb{E}[y]}{\mathbb{E}[f(x)]} f(x), y\right) \), ensuring that \( f(x) \) and \( y \) are compared under the condition of average brightness alignment. At this stage, the GT-mean loss operates under the assumption of consistent image brightness, compelling the model to focus on addressing degradation factors beyond brightness enhancement, such as noise, contrast, or structural details. This dynamic adjustment enables the model to refine its performance progressively, further mitigating degradation while preserving the capabilities of the baseline model.

In application, flexibility is the primary advantage of the GT-mean loss, as it is independent of the model architecture and only requires the enhanced image $f(x)$ and the ground truth $y$ as inputs. Therefore, the GT-mean loss can be used in any supervised LLIE method by simply extending its supervised loss function into the GT-mean version.

The GT-mean loss is also efficient. Compared to the original loss, it doubles the computation, and $W$ is estimated with an analytical solution. Despite there is extra computation overhead, the overall increase introduced by the GT-mean loss is negligible throughout the training process.

\section{Experiment}

\subsection{Datasets and Settings} 
\textbf{Baselines.}
We selected a total of seven supervised LLIE models as baselines. To validate the combined effect of the GT-Mean loss and perceptual loss, the baselines include SNR-aware~\cite{9878461} and CID-Net~\cite{feng2024hvi}, both of which incorporate perceptual loss. All baselines and their original loss functions are listed in Table \ref{table2}:
\begin{table}[h]
	\centering
	\caption{Baselines and Their Loss Functions}
	\resizebox{0.35\textwidth}{!}{
		\begin{tabular}{ll}
			\toprule
			\textbf{Method}       & \textbf{Loss Function} \\\midrule \midrule
			Restormer ~\cite{Zamir2021Restormer}            & $L_1$ loss                \\ 
			RetinexFormer ~\cite{Cai_2023_ICCV}         & $L_1$ loss                \\ 
			LLFormer ~\cite{wang2023ultra}             & Smooth $L_1$ loss ~\cite{Girshick_2015_ICCV}         \\ 
			MIRNet ~\cite{Zamir2020MIRNet}               & Charbonnier loss ~\cite{8954089}       \\ 
			Uformer ~\cite{Wang_2022_CVPR}             & Charbonnier loss~\cite{8954089}        \\ 
			SNR-Aware ~\cite{9878461}  & Charbonnier loss~\cite{8954089}, perceptual loss ~\cite{Johnson_Alahi_FeiFei_2016}\\ 
			CID-Net ~\cite{feng2024hvi}              & $L_1$ loss, edge loss ~\cite{8461664}, perceptual loss~\cite{Johnson_Alahi_FeiFei_2016} \\ 
			\bottomrule
		\end{tabular}
		\label{table2}}
\end{table}
\begin{table*}[h]
    \caption{Comparison on the Paired Datasets. {\color[HTML]{FF0000}Red}{\color[HTML]{00AA00}(Green)} denotes the {\color[HTML]{FF0000}improvement}{\color[HTML]{00AA00}(reduction)} of performance.}
    \label{table3}
    \resizebox{1\textwidth}{!}{
    \begin{tabular}{l|l|ccc|ccc|cc}
      \toprule
       \multicolumn{1}{c|}{\multirow{2}{*}{\textbf{Datasets}}} & \multicolumn{1}{c|}{\multirow{2}{*}{\textbf{Methods}}} & \multicolumn{3}{c|}{\textbf{Normal}}               & \multicolumn{3}{c|}{\textbf{GT-mean}}                                                                                                                                                     & \multicolumn{2}{c}{\textbf{ }} \\
        \multicolumn{1}{c|}{}                             &                                & \textbf{PSNR$\uparrow$}                                                 & \textbf{SSIM$\uparrow$}                                     & \textbf{LPIPS$\downarrow$}                                  & \textbf{PSNR$\uparrow$}                                      & \textbf{SSIM$\uparrow$}                                      & \textbf{LPIPS$\downarrow$}                                  &    \textbf{IQA    $\uparrow$}      &       \textbf{IAA$\uparrow$}                     \\ \midrule \midrule
        \multirow{14}{*}{LOLv1}                                 & Uformer                                                & 18.218                                                                  & 0.771                                                       & 0.212                                                       & 22.325                                                       & 0.81                                                         & 0.195                                                       & 3.317&1.959                                 \\
                                                                & Uformer with GT-mean loss \textbf{(ours)}                & 18.915 \color[HTML]{FF0000}\footnotesize(+0.697)                        & 0.795 \color[HTML]{FF0000}\footnotesize(+0.023)             & 0.168 \color[HTML]{FF0000}\footnotesize(-0.044)             & 22.854 \color[HTML]{FF0000}\footnotesize(+0.529)             & 0.830 \color[HTML]{FF0000}\footnotesize(+0.019)              & 0.157 \color[HTML]{FF0000}\footnotesize(-0.038)             & 3.331\color[HTML]{FF0000}\footnotesize($\uparrow$)&1.971\color[HTML]{FF0000}\footnotesize($\uparrow$)                                 \\ \cmidrule(lr){2-10}
                                                                & MIRNet                                                 & 21.512                                                                  & 0.788                                                       & 0.222                                                       & 24.968                                                       & 0.8                                                          & 0.216                                                       & 2.917&1.745                                 \\
                                                                & MIRNet with GT-mean loss \textbf{(ours)}                 & 21.780 \color[HTML]{FF0000}\footnotesize(+0.268)                        & 0.804 \color[HTML]{FF0000}\footnotesize(+0.016)             & 0.196 \color[HTML]{FF0000}\footnotesize(-0.026)             & 25.596 \color[HTML]{FF0000}\footnotesize(+0.628)             & 0.818 \color[HTML]{FF0000}\footnotesize(+0.018)              & 0.189 \color[HTML]{FF0000}\footnotesize(-0.027)             & 3.039\color[HTML]{FF0000}\footnotesize($\uparrow$)&1.758\color[HTML]{FF0000}\footnotesize($\uparrow$)                                 \\  \cmidrule(lr){2-10}
                                                                & RetinexFormer                                          & 23.83                                                                   & 0.832                                                       & 0.141                                                       & 26.312                                                       & 0.844                                                        & 0.134                                                       & 3.027&1.800                                 \\
                                                                & RetinexFormer with GT-mean loss \textbf{(ours)}          & 24.561 \color[HTML]{FF0000}\footnotesize(+0.731)                        & 0.834 \color[HTML]{FF0000}\footnotesize(+0.003)             & 0.138 \color[HTML]{FF0000}\footnotesize(-0.003)             & 26.586 \color[HTML]{FF0000}\footnotesize(+0.274)             & 0.849 \color[HTML]{FF0000}\footnotesize(+0.005)              & 0.132 \color[HTML]{FF0000}\footnotesize(-0.002)             & 3.373\color[HTML]{FF0000}\footnotesize($\uparrow$)&1.956\color[HTML]{FF0000}\footnotesize($\uparrow$)                                 \\  \cmidrule(lr){2-10}
                                                                & Restormer                                              & 22.718                                                                  & 0.83                                                        & 0.128                                                       & 26.375                                                       & 0.848                                                        & 0.122                                                       & 3.567&2.032                                 \\
                                                                & Restormer with GT-mean loss \textbf{(ours)}              & 23.313 \color[HTML]{FF0000}\footnotesize(+0.595)                        & 0.837 \color[HTML]{FF0000}\footnotesize(+0.007)             & 0.122 \color[HTML]{FF0000}\footnotesize(-0.006)             & 26.743 \color[HTML]{FF0000}\footnotesize(+0.368)             & 0.855 \color[HTML]{FF0000}\footnotesize(+0.007)              & 0.117 \color[HTML]{FF0000}\footnotesize(-0.005)             & 3.672\color[HTML]{FF0000}\footnotesize($\uparrow$)&2.054\color[HTML]{FF0000}\footnotesize($\uparrow$)                                 \\  \cmidrule(lr){2-10}
                                                                & LLFormer                                               & 23.007                                                                  & 0.805                                                       & 0.183                                                       & 25.762                                                       & 0.823                                                        & 0.178                                                       & 3.087&1.946                                 \\
                                                                & LLFormer with GT-mean loss \textbf{(ours)}               & 23.847 \color[HTML]{FF0000}\footnotesize(+0.840)                        & 0.830 \color[HTML]{FF0000}\footnotesize( {+0.025}) & 0.138 \color[HTML]{FF0000}\footnotesize( {-0.045})  & 26.769 \color[HTML]{FF0000}\footnotesize( {+1.007}) & 0.846 \color[HTML]{FF0000}\footnotesize( {+0.023})   & 0.133 \color[HTML]{FF0000}\footnotesize( {-0.045}) & 3.419\color[HTML]{FF0000}\footnotesize($\uparrow$)&2.049\color[HTML]{FF0000}\footnotesize($\uparrow$)                                 \\ \cmidrule(lr){2-10}
                                                                & SNR-Aware                                              & 23.005                                                                  & 0.824                                                       & 0.164                                                       & 26.373                                                       & 0.843                                                        & 0.158                                                       & 3.330&1.893                                 \\
                                                                & SNR-Aware with GT-mean loss \textbf{(ours)}              & 23.992 \color[HTML]{FF0000}\footnotesize(+0.988)                        & 0.836 \color[HTML]{FF0000}\footnotesize(+0.012)             & 0.158 \color[HTML]{FF0000}\footnotesize(-0.006)             & 26.942 \color[HTML]{FF0000}\footnotesize(+0.569)             & 0.853 \color[HTML]{FF0000}\footnotesize(+0.009)              & 0.153 \color[HTML]{FF0000}\footnotesize(-0.005)             & 3.509\color[HTML]{FF0000}\footnotesize($\uparrow$)&1.913\color[HTML]{FF0000}\footnotesize($\uparrow$)                                 \\  \cmidrule(lr){2-10}
                                                                & CID-Net                                                & 23.809                                                                  & 0.857                                                       & 0.086                                                       & 27.715                                                       & 0.876                                                        & 0.079                                                       & 4.087&2.157                                 \\
                                                                & CID-Net with GT-mean loss \textbf{(ours)}                & {25.122} \color[HTML]{FF0000}\footnotesize( {+1.313}) & {0.865} \color[HTML]{FF0000}\footnotesize(+0.008)  & {0.081} \color[HTML]{FF0000}\footnotesize(-0.005)  & {28.108} \color[HTML]{FF0000}\footnotesize(+0.393)   & {0.878} \color[HTML]{FF0000}\footnotesize(+0.002)   & {0.075} \color[HTML]{FF0000}\footnotesize(-0.004)  & 4.074\color[HTML]{00AA00}\footnotesize($\downarrow$)&2.161\color[HTML]{FF0000}\footnotesize($\uparrow$)                                 \\ \midrule  \midrule
        \multirow{12}{*}{LOLv2-real}                            & Uformer                                                & 14.941                                                                  & 0.76                                                        & 0.228                                                       & 22.148                                                       & 0.831                                                        & 0.199                                                       & 3.478&2.009                                 \\
                                                                & Uformer with GT-mean loss \textbf{(ours)}                & 16.103 \color[HTML]{FF0000}\footnotesize( {+1.162})            & 0.792 \color[HTML]{FF0000}\footnotesize(+0.032)             & 0.180 \color[HTML]{FF0000}\footnotesize(-0.048)             & 23.989 \color[HTML]{FF0000}\footnotesize( {+1.841}) & 0.858 \color[HTML]{FF0000}\footnotesize(+0.026)              & 0.156 \color[HTML]{FF0000}\footnotesize(-0.043)             & 3.778\color[HTML]{FF0000}\footnotesize($\uparrow$)&2.048\color[HTML]{FF0000}\footnotesize($\uparrow$)                                 \\  \cmidrule(lr){2-10}
                                                                & MIRNet                                                 & 21.648                                                                  & 0.81                                                        & 0.313                                                       & 26.712                                                       & 0.827                                                        & 0.303                                                       & 2.598&1.520                                 \\
                                                                & MIRNet with GT-mean loss \textbf{(ours)}                 & 22.050 \color[HTML]{FF0000}\footnotesize(+0.402)                        & 0.830 \color[HTML]{FF0000}\footnotesize(+0.021)             & 0.214 \color[HTML]{FF0000}\footnotesize( {-0.099})  & 26.769 \color[HTML]{FF0000}\footnotesize(+0.057)             & 0.846 \color[HTML]{FF0000}\footnotesize(+0.019)              & 0.208 \color[HTML]{FF0000}\footnotesize( {-0.101}) & 2.924\color[HTML]{FF0000}\footnotesize($\uparrow$)&1.702\color[HTML]{FF0000}\footnotesize($\uparrow$)                                 \\  \cmidrule(lr){2-10}
                                                                & RetinexFormer                                          & 21.272                                                                  & 0.841                                                       & 0.163                                                       & 27.65                                                        & 0.877                                                        & 0.152                                                       & 2.714&1.590                                 \\
                                                                & RetinexFormer with GT-mean loss \textbf{(ours)}          & 21.810 \color[HTML]{FF0000}\footnotesize(+0.538)                        & {0.852} \color[HTML]{FF0000}\footnotesize(+0.011)  & {0.143} \color[HTML]{FF0000}\footnotesize(-0.020)  & 28.437 \color[HTML]{FF0000}\footnotesize(+0.787)             & 0.879 \color[HTML]{FF0000}\footnotesize(+0.002)              & {0.134} \color[HTML]{FF0000}\footnotesize(-0.018)  & 3.206\color[HTML]{FF0000}\footnotesize($\uparrow$)&1.884\color[HTML]{FF0000}\footnotesize($\uparrow$)                                 \\ \cmidrule(lr){2-10}
                                                                & Restormer                                              & 20.235                                                                  & 0.841                                                       & 0.162                                                       & 28.159                                                       & 0.88                                                         & 0.147                                                       & 3.478&1.987                                 \\
                                                                & Restormer with GT-mean loss \textbf{(ours)}              & 20.717 \color[HTML]{FF0000}\footnotesize(+0.482)                        & 0.845 \color[HTML]{FF0000}\footnotesize(+0.004)             & 0.149 \color[HTML]{FF0000}\footnotesize(-0.013)             & {28.440} \color[HTML]{FF0000}\footnotesize(+0.281)  & {0.884} \color[HTML]{FF0000}\footnotesize(+0.004)   & 0.135 \color[HTML]{FF0000}\footnotesize(-0.012)             & 3.554\color[HTML]{FF0000}\footnotesize($\uparrow$)&2.020\color[HTML]{FF0000}\footnotesize($\uparrow$)                                 \\  \cmidrule(lr){2-10}
                                                                & LLFormer                                               & 21.308                                                                  & 0.803                                                       & 0.248                                                       & 27.052                                                       & 0.828                                                        & 0.236                                                       & 2.882&1.827                                 \\
                                                                & LLFormer with GT-mean loss \textbf{(ours)}               & {22.291} \color[HTML]{FF0000}\footnotesize(+0.983)             & 0.844 \color[HTML]{FF0000}\footnotesize( {+0.041}) & 0.166 \color[HTML]{FF0000}\footnotesize(-0.082)             & 28.334 \color[HTML]{FF0000}\footnotesize(+1.282)             & 0.870 \color[HTML]{FF0000}\footnotesize( {+0.042})   & 0.156 \color[HTML]{FF0000}\footnotesize(-0.080)             & 3.104\color[HTML]{FF0000}\footnotesize($\uparrow$)&1.880\color[HTML]{FF0000}\footnotesize($\uparrow$)                                 \\ \cmidrule(lr){2-10}
                                                                & SNR-Aware                                              & 21.103                                                                  & 0.839                                                       & 0.169                                                                & 26.971                                                       & 0.866                                                        & 0.161                                                       & 3.354&1.879                                 \\
                                                                & SNR-Aware with GT-mean loss \textbf{(ours)}              & 21.350 \color[HTML]{FF0000}\footnotesize(+0.247)                        & 0.844 \color[HTML]{FF0000}\footnotesize(+0.005)             & 0.164 \color[HTML]{FF0000}\footnotesize(-0.005)                     & 27.740 \color[HTML]{FF0000}\footnotesize(+0.770)             & 0.875 \color[HTML]{FF0000}\footnotesize(+0.010)              & 0.154 \color[HTML]{FF0000}\footnotesize(-0.007)             & 3.468\color[HTML]{FF0000}\footnotesize($\uparrow$)&1.889\color[HTML]{FF0000}\footnotesize($\uparrow$)                                 \\ \midrule \midrule
        \multirow{12}{*}{LOLv2-synthetic}                       & Uformer                                                & 24.693                                                                  & 0.932                                                     & 0.060                                                                   & 27.438                                                       & 0.941                                                       & 0.055                                                      & 3.148&2.114                                 \\
                                                                & Uformer with GT-mean loss \textbf{(ours)}                & 25.319 \color[HTML]{FF0000}\footnotesize(+0.626)                        & 0.940 \color[HTML]{FF0000}\footnotesize( +0.007) & 0.049 \color[HTML]{FF0000}\footnotesize(-0.011)                               & 28.683\color[HTML]{FF0000}\footnotesize(+1.245)              & 0.948\color[HTML]{FF0000}\footnotesize(+0.007) & 0.045 \color[HTML]{FF0000}\footnotesize(-0.010)            & 3.191\color[HTML]{FF0000}\footnotesize($\uparrow$)&2.144\color[HTML]{FF0000}\footnotesize($\uparrow$)                                 \\  \cmidrule(lr){2-10}
                                                                & MIRNet                                                 & 22.059                                                                  & 0.894                                                     & 0.122                                                                  & 25.274                                                        & 0.908                                                       & 0.114                                                      & 2.956&2.145                                 \\
                                                                & MIRNet with GT-mean loss \textbf{(ours)}                 & 22.576 \color[HTML]{FF0000}\footnotesize(+0.517)                        & 0.906 \color[HTML]{FF0000}\footnotesize(+0.011)            & 0.104 \color[HTML]{FF0000}\footnotesize({-0.018})             &26.215\color[HTML]{FF0000}\footnotesize(+0.941)   & 0.918\color[HTML]{FF0000}\footnotesize(+0.010)            & 0.094\color[HTML]{FF0000}\footnotesize({-0.020})           & 3.064\color[HTML]{FF0000}\footnotesize($\uparrow$)&2.187\color[HTML]{FF0000}\footnotesize($\uparrow$)                                 \\  \cmidrule(lr){2-10}
                                                                & RetinexFormer                                          & 25.281                                                                  & 0.928                                                      & 0.064                                                                 &28.827                                                        & 0.939                                                      &0.057                                                      & 3.102&2.099                                 \\
                                                                & RetinexFormer with GT-mean loss \textbf{(ours)}          & 25.583 \color[HTML]{FF0000}\footnotesize(+0.299)                        & 0.933 \color[HTML]{FF0000}\footnotesize(+0.005)            & 0.063 \color[HTML]{FF0000}\footnotesize(-0.001)                     & 29.261\color[HTML]{FF0000}\footnotesize(+0.434)   & 0.944\color[HTML]{FF0000}\footnotesize(+0.005)             & 0.056\color[HTML]{FF0000}\footnotesize(-0.001)            & 3.197\color[HTML]{FF0000}\footnotesize($\uparrow$)&2.130\color[HTML]{FF0000}\footnotesize($\uparrow$)                                 \\ \cmidrule(lr){2-10}
                                                                & Restormer                                              & 26.288                                                                  & 0.944                                                      & 0.045                                                                 & 30.57                                                       &0.955                                                    & 0.039                                                        & 3.350&2.187                                 \\
                                                                & Restormer with GT-mean loss \textbf{(ours)}              & {26.630} \color[HTML]{FF0000}\footnotesize(+0.342)             & {0.946} \color[HTML]{FF0000}\footnotesize(+0.002)    & {0.041} \color[HTML]{FF0000}\footnotesize(-0.004)            & {31.001}\color[HTML]{FF0000}\footnotesize(+0.431)              & {0.957}\color[HTML]{FF0000}\footnotesize(+0.002)  & {0.036} \color[HTML]{FF0000}\footnotesize(-0.003)  & 3.404\color[HTML]{FF0000}\footnotesize($\uparrow$)&2.218\color[HTML]{FF0000}\footnotesize($\uparrow$)                                 \\  \cmidrule(lr){2-10}
                                                                & LLFormer                                               & 24.195                                                                  & 0.918                                                      & 0.07                                                                 & 27.862                                                        & 0.930                                                       & 0.064                                                     & 3.176&2.137                                 \\
                                                                & LLFormer with GT-mean loss \textbf{(ours)}               & 25.152 \color[HTML]{FF0000}\footnotesize( {+0.957})            & 0.932 \color[HTML]{FF0000}\footnotesize({+0.014}) & 0.058 \color[HTML]{FF0000}\footnotesize(-0.012)                         & 29.266\color[HTML]{FF0000}\footnotesize({+1.404})             & 0.945\color[HTML]{FF0000}\footnotesize({+0.015})             & 0.051 \color[HTML]{FF0000}\footnotesize(-0.013) & 3.283\color[HTML]{FF0000}\footnotesize($\uparrow$)&2.177\color[HTML]{FF0000}\footnotesize($\uparrow$)                                 \\  \cmidrule(lr){2-10}
                                                                & SNR-Aware                                              & 24.173                                                                  & 0.924                                                      & 0.064                                                                 & 27.756                                                      & 0.937                                                       & 0.058                                                    & 3.275&2.209                                 \\
                                                                & SNR-Aware with GT-mean loss \textbf{(ours)}              & 24.301 \color[HTML]{FF0000}\footnotesize(+0.128)                        & 0.933\color[HTML]{FF0000}\footnotesize(+0.009)            & 0.057\color[HTML]{FF0000}\footnotesize(-0.007)                       & 28.525\color[HTML]{FF0000}\footnotesize(+0.769)              & 0.945\color[HTML]{FF0000}\footnotesize(+0.008)             & 0.050 \color[HTML]{FF0000}\footnotesize(-0.008)             & 3.326\color[HTML]{FF0000}\footnotesize($\uparrow$)&2.207\color[HTML]{00AA00}\footnotesize($\downarrow$)                                \\ \bottomrule
        \end{tabular}}
    \end{table*}\\
\textbf{Datasets.} We conducted experiments on both paired and unpaired datasets to evaluate our loss. For paired datasets, we used LOLv1~\cite{Chen_Wang_Yang_Liu_2018}, LOLv2-real ~\cite{Yang_Wang_Huang_Wang_Liu_2021}, and LOLv2-syn ~\cite{Yang_Wang_Huang_Wang_Liu_2021}.  Specifically, the LOLv1 dataset includes 485 training images and 15 testing images. The LOLv2-real dataset includes 689 training images and 100 testing images. The LOLv2-synthetic dataset includes 900 training images and 100 testing images. For unpaired datasets, we chose DICM ~\cite{b6ChulwooLee2013ContrastEB}, VV ~\cite{Vonikakis_Kouskouridas_Gasteratos_2018}, NPE ~\cite{NPE}, MEF ~\cite{MEF}, and LIME ~\cite{LIME} as the test sets. \\
\textbf{Evaluation Metrics.} For the paired datasets, we employed standard evaluation metrics, including PSNR, SSIM \cite{Wang_Bovik_Sheikh_Simoncelli_2004}, and LPIPS \cite{8578166}. To further align with human perception, we utilized the large multi-modality model QALIGN~\cite{wu24ah}, specifically its image quality assessment (IQA) and image aesthetic assessment (IAA) modules, for more perceptually meaningful evaluations.

For unpaired datasets, we evaluated the model performance using QALIGN's IQA and IAA modules, NIQE~\cite{Mittal_Soundararajan_Bovik_2013}, and MUSIQ~\cite{ke2021musiq}.\\
\textbf{Implementation Details.}
To retrain the baselines equipped with GT-mean loss functions, we followed the their official settings, which can be found in Appendix.

\subsection{Quantitative Results}
\label{4_2}
\textbf{Paired Datasets.}
As shown in Table \ref{table3}, we evaluated the performance of the GT-Mean loss on three paired datasets, covering reference-based evaluation metrics (PSNR, SSIM, LPIPS) and their GT-Mean versions (GT-Mean PSNR, GT-Mean SSIM, GT-Mean LPIPS), along with QALIGN for image quality assessment (IQA) and image alignment assessment (IAA).

The experimental results demonstrate that, across the seven baseline models, regardless of the type of original loss function, extending them with the GT-Mean loss consistently improves performance across all evaluation metrics. This outcome validates the universal effectiveness of the GT-Mean loss in LLIE tasks.
Notably, both SNR-aware and CIDNet methods exhibit significant performance improvements after incorporating the GT-Mean loss, indicating that the GT-Mean loss effectively complements perceptual loss. One possible explanation is that the GT-Mean loss compensates for the lack of pixel-level perception in perceptual loss. 
Furthermore, since the weight $W$ is computed analytically during training and does not require gradient propagation, its cost is negligible. Compared to the model's forward and backward propagation, the original loss computation is minimal. The second term in the GT-Mean Loss is essentially a brightness-adjusted version of the original loss, sharing the same computational complexity. Thus, the GT-Mean Loss incurs only twice the computational overhead of the original loss. For example, using GT-Mean Loss with Retinexformer increases training time by just 1\%. Therefore, this loss function can be seamlessly integrated into existing supervised LLIE methods, achieving nearly zero-cost performance improvements.\\
\begin{figure}[H]
\captionsetup[subfigure]{labelformat=empty}
    
 \begin{subfigure}[b]{0.32\linewidth} 
        \centering
        \includegraphics[width=\linewidth]{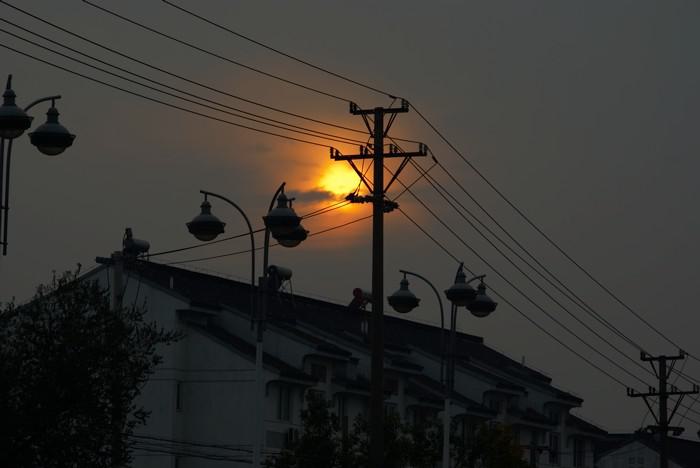}
        \caption{Input}
    \end{subfigure}
    \begin{subfigure}[b]{0.32\linewidth} 
        \centering
        \includegraphics[width=\linewidth]{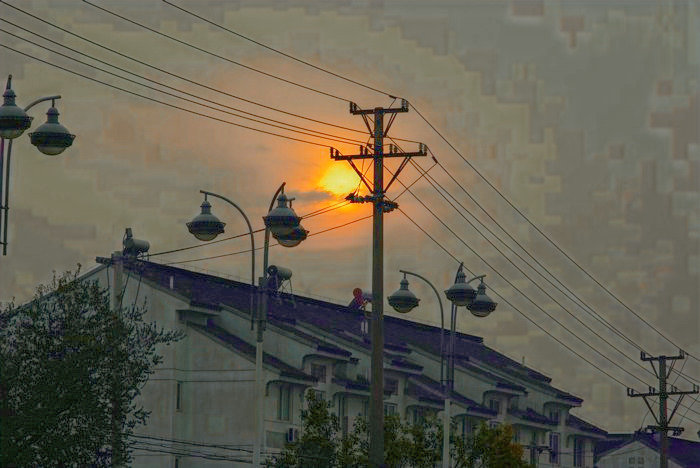}
        \caption{RetinexFormer}
    \end{subfigure}
    \begin{subfigure}[b]{0.32\linewidth} 
        \centering
        \includegraphics[width=\linewidth]{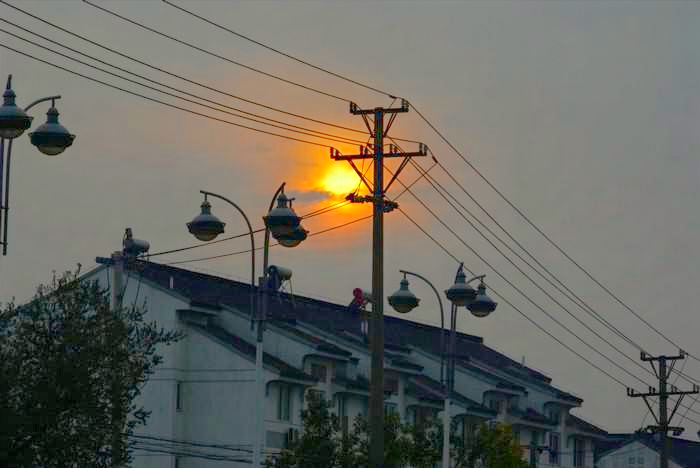}
        \caption{Ours}
    \end{subfigure}

 \begin{subfigure}[b]{0.32\linewidth} 
        \centering
        \includegraphics[width=\linewidth]{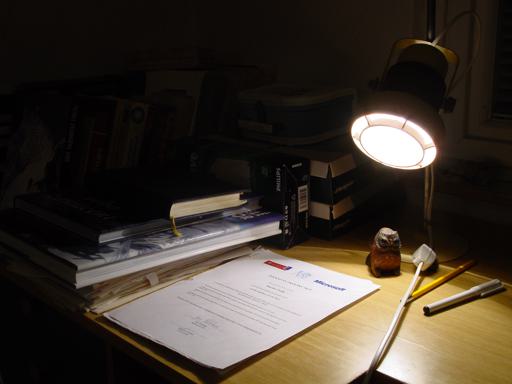}
        \caption{Input}
    \end{subfigure}
    \begin{subfigure}[b]{0.32\linewidth} 
        \centering
        \includegraphics[width=\linewidth]{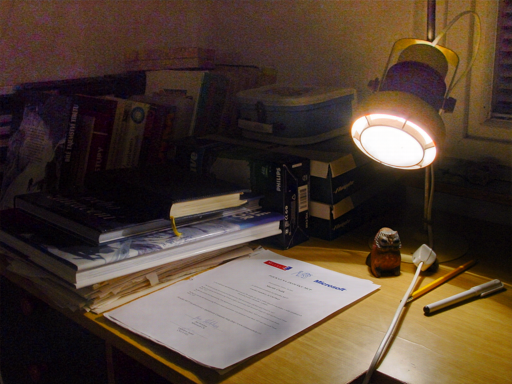}
        \caption{Uformer}
    \end{subfigure}
    \begin{subfigure}[b]{0.32\linewidth} 
        \centering
        \includegraphics[width=\linewidth]{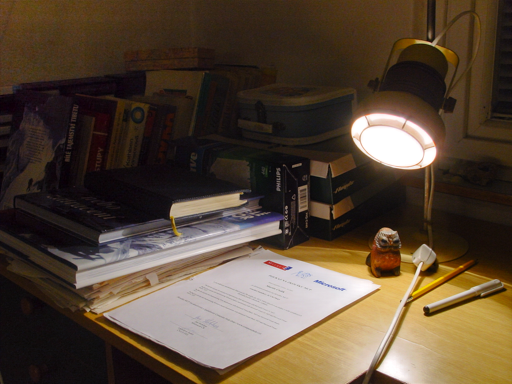}
        \caption{Ours}
\end{subfigure}

\begin{subfigure}[b]{0.32\linewidth} 
        \centering
        \includegraphics[width=\linewidth]{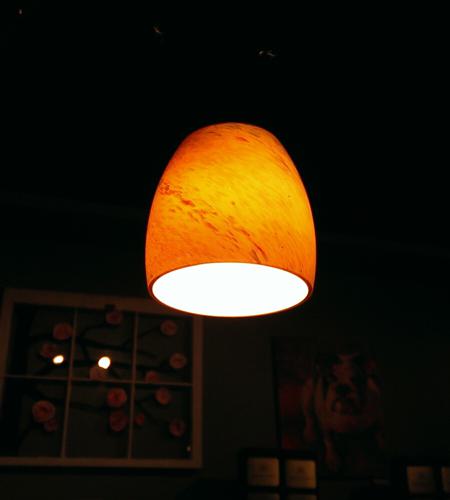}
        \caption{Input}
    \end{subfigure}
    \begin{subfigure}[b]{0.32\linewidth} 
        \centering
        \includegraphics[width=\linewidth]{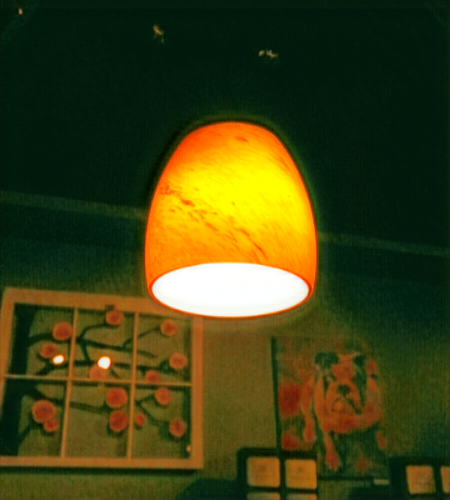}
        \caption{MIRNet}
    \end{subfigure}
    \begin{subfigure}[b]{0.32\linewidth} 
        \centering
        \includegraphics[width=\linewidth]{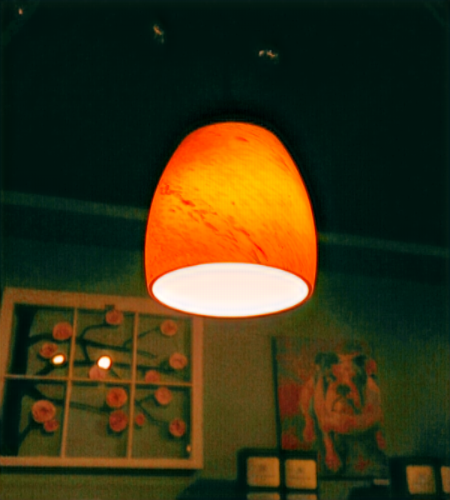}
        \caption{Ours}
    \end{subfigure}

 \begin{subfigure}[b]{0.32\linewidth} 
        \centering
        \includegraphics[width=\linewidth]{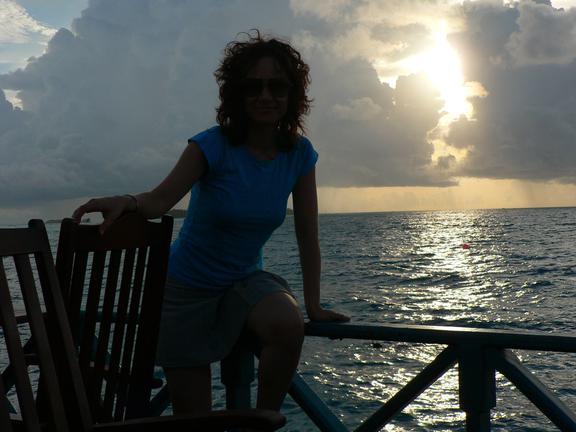}
        \caption{Input}
    \end{subfigure}
    \begin{subfigure}[b]{0.32\linewidth} 
        \centering
        \includegraphics[width=\linewidth]{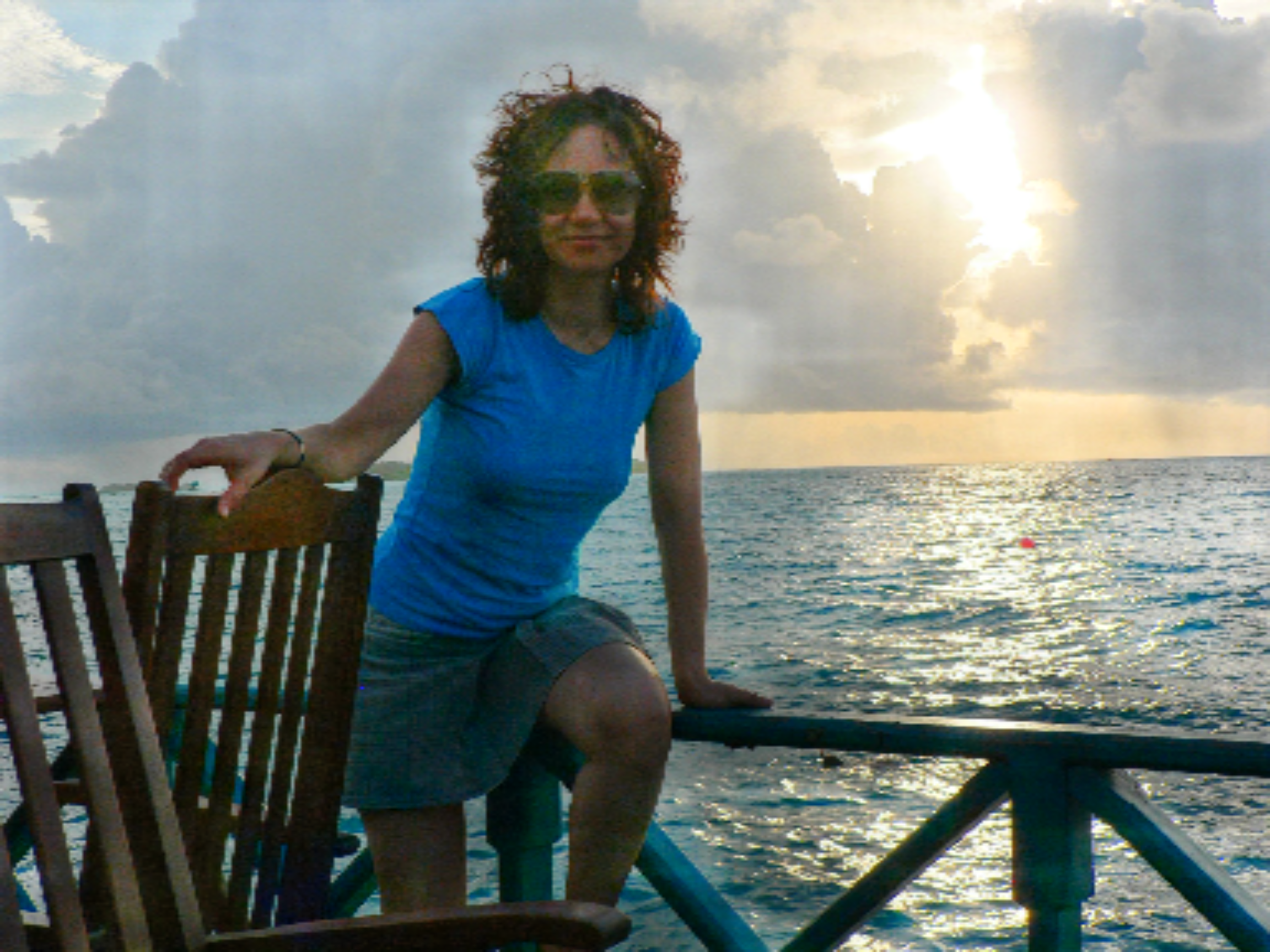}
        \caption{SNR-aware}
    \end{subfigure}
    \begin{subfigure}[b]{0.32\linewidth} 
        \centering
        \includegraphics[width=\linewidth]{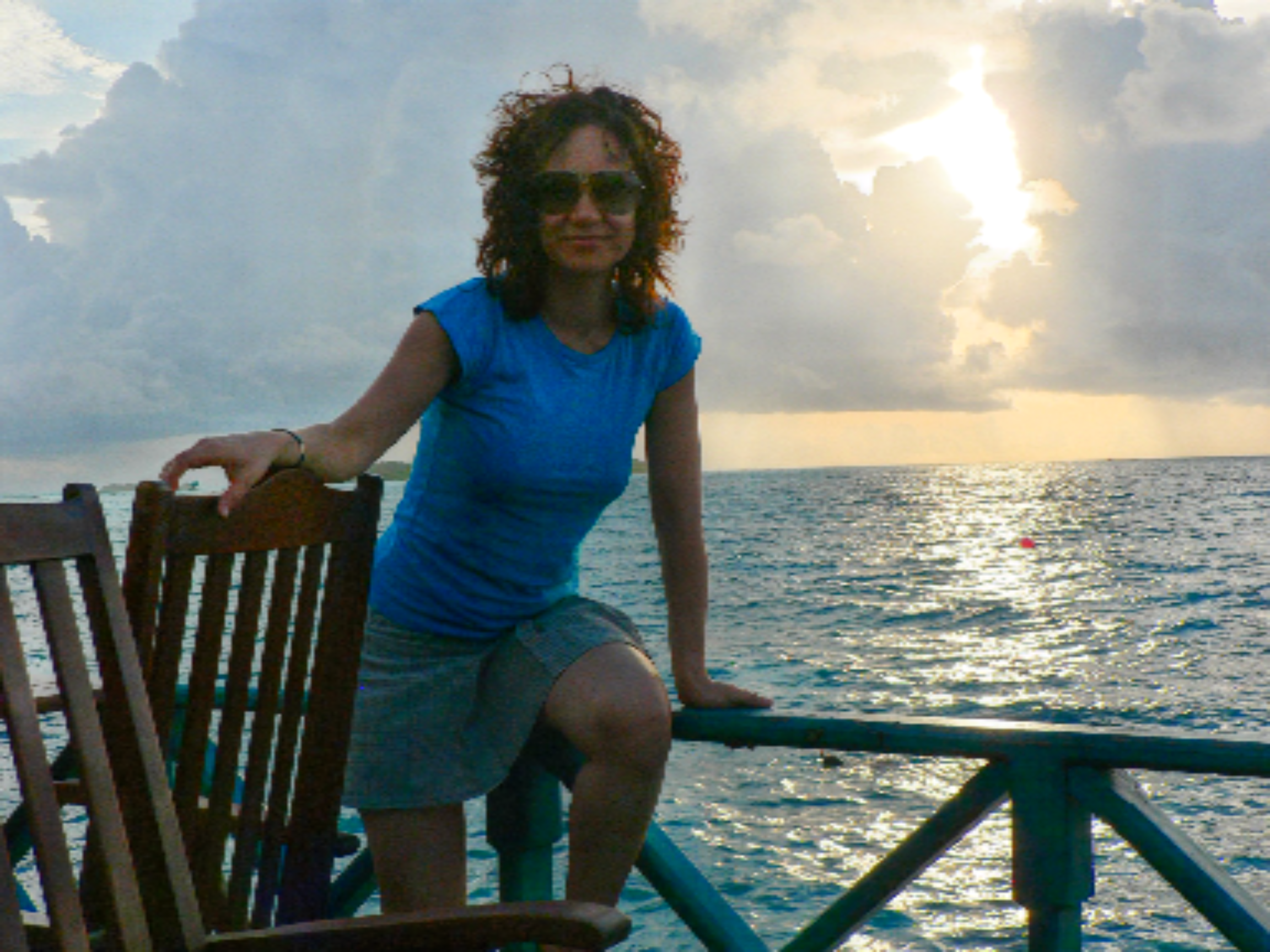}
        \caption{Ours}
    \end{subfigure}

 \begin{subfigure}[b]{0.32\linewidth} 
        \centering
        \includegraphics[width=\linewidth]{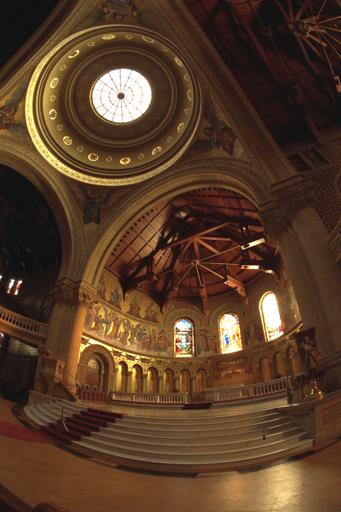}
        \caption{Input}
    \end{subfigure}
    \begin{subfigure}[b]{0.32\linewidth} 
        \centering
        \includegraphics[width=\linewidth]{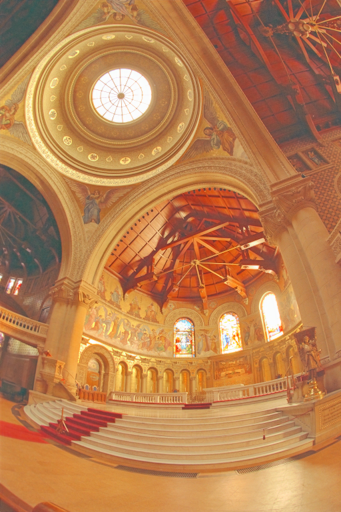}
        \caption{Restormer}
    \end{subfigure}
    \begin{subfigure}[b]{0.32\linewidth} 
        \centering
        \includegraphics[width=\linewidth]{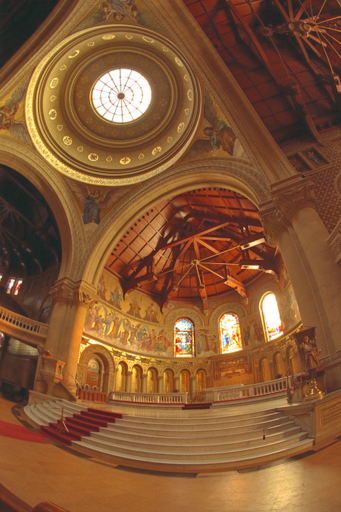}
        \caption{Ours}
\end{subfigure}

 \begin{subfigure}[b]{0.32\linewidth} 
        \centering
        \includegraphics[width=\linewidth]{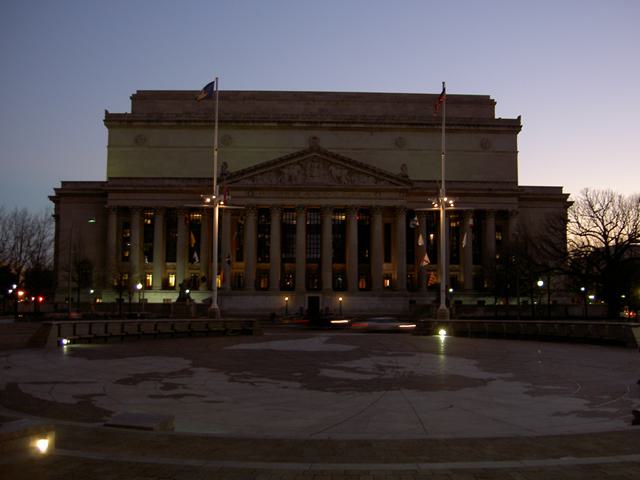}
        \caption{Input}
    \end{subfigure}
    \begin{subfigure}[b]{0.32\linewidth} 
        \centering
        \includegraphics[width=\linewidth]{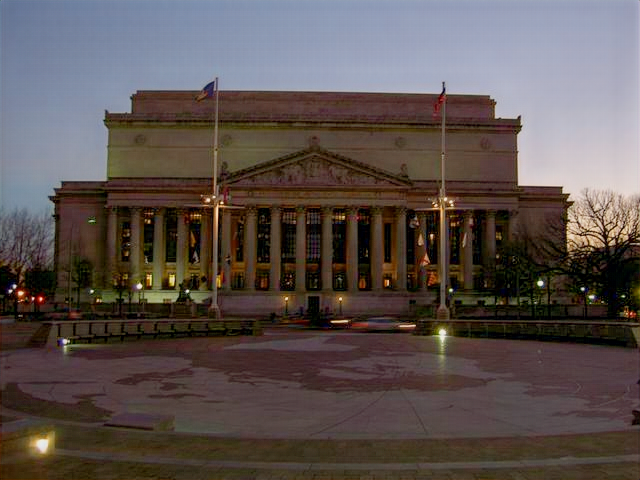}
        \caption{LLformer}
    \end{subfigure}
    \begin{subfigure}[b]{0.32\linewidth} 
        \centering
        \includegraphics[width=\linewidth]{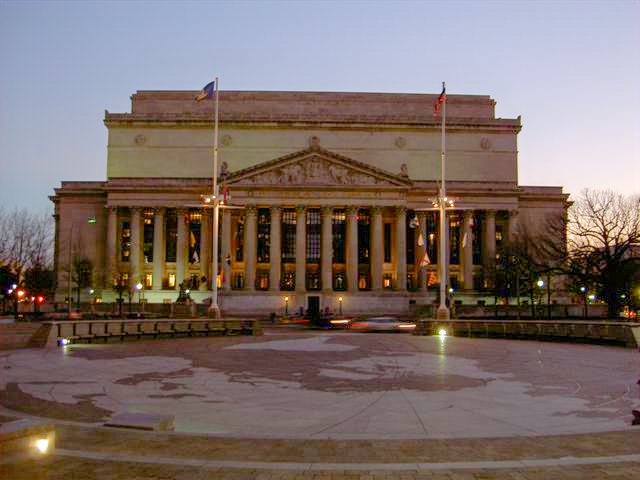}
        \caption{Ours}
\end{subfigure}
\caption{Visual comparison of results on the five unpaired datasets: original image (left), results from baselines (middle), and results from their improved versions based on our loss (right). }   
\label{fiveset}
\end{figure} 

\begin{table*}[htb]
    \centering
    
    \caption{Comparison on the Unpaired Datasets. {\color[HTML]{FF0000}Red}{\color[HTML]{00AA00}(Green)}  denotes the {\color[HTML]{FF0000}improvement}{\color[HTML]{00AA00}(reduction)} of performance. Note that all the models were trained on the LOLv2-synthetic dataset.}
    \label{table4}
    \resizebox{1\textwidth}{!}{
        \begin{tabular}{l|cccc|cccc|cccc|cccc|cccc|cccc}
       \hline
       \multicolumn{1}{c|}{\multirow{2}{*}{Method}}    & \multicolumn{4}{c|}{DICM}                                             & \multicolumn{4}{c|}{MEF}                                             & \multicolumn{4}{c|}{LIME}                                            & \multicolumn{4}{c|}{NPE}                                             &\multicolumn{4}{c|}{VV}   &          \multicolumn{4}{c}{AVG}      \\
                                 & NIQE$\downarrow$ & MUSIQ $\uparrow$ & IQA $\uparrow$ & IAA $\uparrow$ & NIQE$\downarrow$ & MUSIQ$\uparrow$ & IQA $\uparrow$ & IAA $\uparrow$ & NIQE$\downarrow$ & MUSIQ$\uparrow$ & IQA $\uparrow$ & IAA $\uparrow$ & NIQE$\downarrow$ & MUSIQ$\uparrow$ & IQA $\uparrow$ & IAA $\uparrow$ & NIQE$\downarrow$ & MUSIQ$\uparrow$ & IQA $\uparrow$ & IAA $\uparrow$ & NIQE$\downarrow$                           & MUSIQ$\uparrow$                            & IQA $\uparrow$ & IAA $\uparrow$ \\ \midrule
       Restormer                                       & 3.220            & 58.525           & 3.885          & 2.800          & 3.660            & 56.528          & 3.215         & 2.415           & 3.660            & 58.461          & 3.267          & 2.466          & 3.470            & 61.031          & 3.781          & 2.735          & 3.290            & 37.919          & 3.710          & 2.264          & 3.460                                      & 54.493                                     & 3.572          & 2.536          \\
       Restormer with GT-mean loss (\textbf{ours})     & 3.180            & 58.604           & 3.913          & 2.821          & 3.630            & 56.522          & 3.208          &2.431         & 3.630            & 58.124          & 3.380          & 2.521          & 3.450            & 60.971          & 3.820          & 2.769          & 3.300            & 38.290          & 3.712          & 2.255          & 3.440 \color[HTML]{FF0000}\footnotesize(-) & 54.502\color[HTML]{FF0000}\footnotesize(+) & 3.607\color[HTML]{FF0000}\footnotesize(+)          & 2.559\color[HTML]{FF0000}\footnotesize(+)          \\ \midrule
       MIRNET                                          & 3.820            & 52.467           & 3.111          & 2.337          & 3.670            & 47.399          & 2.742         & 2.056          & 4.230            & 54.837          & 2.860          & 2.088          & 3.470            & 58.641          & 3.285          & 2.374          & 3.640            & 54.566          & 2.955          & 2.162          & 3.770                                      & 53.582                                     & 2.991          & 2.203          \\
       MIRNET with GT-mean loss \textbf{(ours)}        & 3.200            & 53.188           & 3.295          & 2.375          & 3.600            & 47.611          & 2.817          & 2.067          & 4.330            & 55.776          & 2.747          & 2.058          & 3.500            & 58.718          & 3.366          & 2.428          & 3.710            & 54.891          & 3.120          & 2.215          & 3.670\color[HTML]{FF0000}\footnotesize(-)  & 54.037\color[HTML]{FF0000}\footnotesize(+) & 3.069\color[HTML]{FF0000}\footnotesize(+)          & 2.229\color[HTML]{FF0000}\footnotesize(+)          \\ \midrule
       Retinexformer                                   & 3.230            & 57.398           & 3.800          & 2.740          & 3.860            & 56.170          & 3.137 & 2.373         & 3.880            & 57.262          & 3.111          & 2.323          & 3.380            & 60.507          & 3.673          & 2.699          & 2.730            & 37.513          & 3.471          & 2.154          & 3.420                                      & 53.77                                      & 3.438          & 2.458          \\
       Retinexformer with GT-mean loss (\textbf{ours}) & 3.210            & 57.247           & 3.805          & 2.773          & 3.820            & 56.633          & 3.195          & 2.412           & 3.840            & 57.374          & 3.273          & 2.423          & 3.370            & 60.682          & 3.706          & 2.719          & 2.770            & 37.654          & 3.517          & 2.166          & 3.400\color[HTML]{FF0000}\footnotesize(-)  & 53.918\color[HTML]{FF0000}\footnotesize(+) & 3.499\color[HTML]{FF0000}\footnotesize(+)          & 2.498\color[HTML]{FF0000}\footnotesize(+)          \\ \midrule
       SNR                                             & 6.070            & 47.025           & 3.379          & 2.526          & 4.270            & 48.685          & 2.531          & 1.943          & 6.060            & 49.216          & 2.836          & 2.102          & 6.470            & 46.441          & 3.445          & 2.551          & 11.520           & 23.186          & 3.067          & 1.955          & 6.880                                      & 42.911                                     & 3.052           & 2.215          \\
       SNR with GT-mean loss (\textbf{ours})           & 6.120            & 47.430           & 3.521          & 2.571          & 4.260            & 48.780          & 2.571         & 1.957          & 6.110            & 49.008          & 2.946          & 2.152          & 6.460            & 46.602          & 3.580          & 2.605          & 11.550           & 23.853          & 3.137          & 1.997          & 6.900\color[HTML]{00AA00}\footnotesize(+)  & 43.135\color[HTML]{FF0000}\footnotesize(+) & 3.151 \color[HTML]{FF0000}\footnotesize(+)          & 2.256 \color[HTML]{FF0000}\footnotesize(+)          \\ \midrule
       Uformer                                         & 3.080            & 58.084           & 2.971          & 2.144          & 3.720            & 56.177          & 3.177          & 2.403          & 3.660            & 57.698          & 2.646          & 1.967          & 3.400            & 61.310          & 2.938          & 2.131          & 2.700            & 36.235          & 2.904          & 1.839          & 3.310                                      & 53.901                                     & 2.927           & 2.097           \\
       Uformer with GT-mean loss (\textbf{ours})       & 3.120            & 58.981           & 3.067          & 2.180          & 3.690            & 56.641          & 3.229           & 2.426          & 3.640            & 58.200          & 2.712          & 1.973          & 3.380            & 61.704          & 2.956          & 2.113          & 2.700            & 36.695          & 3.001          & 1.848          & 3.300\color[HTML]{FF0000}\footnotesize(-)  & 54.444\color[HTML]{FF0000}\footnotesize(+) & 2.993 \color[HTML]{FF0000}\footnotesize(+)         & 2.108 \color[HTML]{FF0000}\footnotesize(+)          \\ \midrule
       LLformer                                        & 3.260            & 56.642           & 3.832          & 2.788          & 3.750            & 53.335          & 2.667          & 1.993          & 4.010            & 55.671          & 3.040          & 2.343          & 3.320            & 59.824          & 3.657          & 2.716          & 3.160            & 60.885          & 3.557          & 2.249          & 3.500                                      & 57.271                                     & 3.351           & 2.418           \\
       LLformer with GT-mean loss \textbf{(ours)}      & 3.050            & 57.038           & 3.910          & 2.837          & 3.650            & 53.842          &2.706          & 2.017          & 4.070            & 55.830          & 3.118          & 2.416          & 3.330            & 60.044          & 3.707          & 2.731          & 2.990            & 60.858          & 3.563          & 2.231          & 3.410\color[HTML]{FF0000}\footnotesize(-)  & 57.522\color[HTML]{FF0000}\footnotesize(+) & 3.401 \color[HTML]{FF0000}\footnotesize(+)          & 2.446 \color[HTML]{FF0000}\footnotesize(+)          \\ \midrule
       \end{tabular}
    }
    \end{table*}
\textbf{Unpaired Datasets.} Table \ref{table4} presents the performance of the models on five unpaired datasets. Compared with the baseline performance, using GT-mean loss demonstrates superior or comparable results in most cases in terms of the non-reference evaluation metrics. These results indicate that the GT-Mean loss possesses strong generalization capabilities: even when tested on unseen images, performance improvements are still observed, proving that its success on supervised datasets is not due to overfitting. As shown in Figure~\ref{fiveset}, the examples based on the GT-Mean loss demonstrate improvements over the baseline results. Specifically, the baseline results of Retinexformer and Uformer show artifacts in sky regions and paper textures respectively. For MIRNet and SNR-Aware, the baseline outputs exhibit noticeable halo artifacts around object contours. Restormer and LLformer represent typical cases of over-exposure and under-exposure in their baseline results. Despite the diverse failure patterns observed across these baseline models, all these issues are effectively mitigated through the integration of our GT-Mean loss. This comprehensive improvement demonstrates the robustness and adaptability of our approach in enhancing image quality across diverse model architectures.

\begin{figure*}[h]
\centering
\includegraphics[width=1\textwidth]{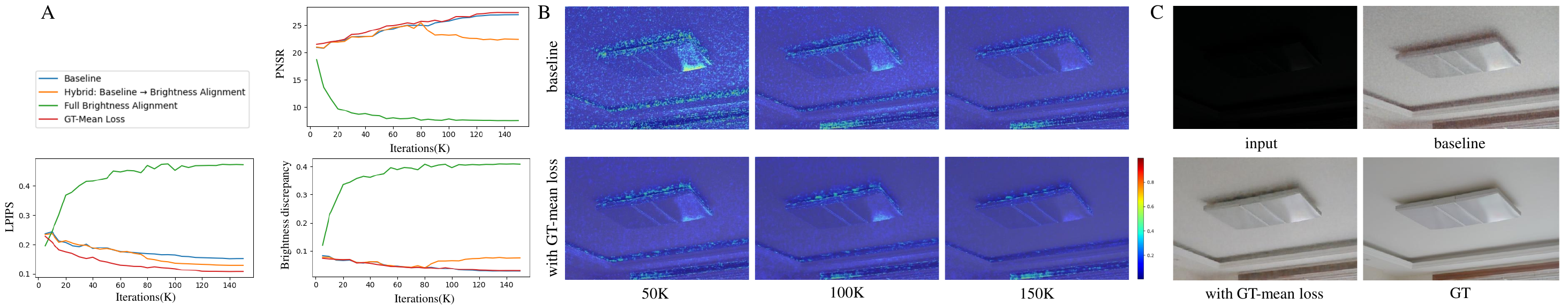} 
    
\caption{Impact of GT-Mean Loss integration on training dynamics and enhancement quality. 
    (A) Evolution of LPIPS (perceptual quality), PSNR (pixel fidelity), and brightness discrepancy (\(|\mathbb{E}[y] - \mathbb{E}[f(x)]|\)) across four training strategies. 
    (B) Absolute residual heatmaps (\(|y - f(x)|\)) at critical training stages. 
    (C) Final output comparison showing improvements in color fidelity and detail preservation.}
\label{fig4}
\end{figure*}
\subsection{Impact of GT-Mean Loss on Training}

\subsubsection{Experimental Configuration.}
We evaluate training dynamics using Retinexformer \cite{Cai_2023_ICCV} on the LOLv1 training dataset with four distinct strategies:  
\begin{itemize}
    \item \textit{Baseline}: Original Retinexformer loss (L1 loss).
    \item \textit{Hybrid}: A phased approach transitioning from the baseline loss (0--70K iterations) to a brightness-aligned loss in Eq.~\ref{eq1} (80K--150K iterations), with linear interpolation between 70K and 80K iterations.
    \item \textit{Full Brightness Alignment}: Exclusive use of the brightness-aligned loss throughout training.
    \item \textit{GT-Mean Loss}: Our proposed dynamic combination of the baseline and brightness-aligned losses.
\end{itemize}

Metrics were recorded every 5K iterations over a total of 150K iterations, including perceptual quality (LPIPS), pixel fidelity (PSNR), and brightness discrepancy \(|\mathbb{E}[y] - \mathbb{E}[f(x)]|\).

\subsubsection{Quantitative Analysis.}  

As shown in Figure~\ref{fig4}A, three key findings are revealed:\\
\textbf{Does brightness alignment improve perception?} We first compare the hybrid strategy with baseline. When brightness alignment is introduced at 80K iterations, the hybrid approach achieves 0.129 LPIPS (vs. baseline 0.155), indicating enhanced perceptual quality. However, this comes at the cost of increased mean brightness discrepancy (from 0.028 to 0.075) and  4.6dB PSNR drop. These results suggest that while the introduction of brightness alignment in the hybrid strategy benefits human perception, its linear implementation disrupts the learning of brightness distribution.\\
\textbf{Can brightness alignment alone achieve performance improvement?} We found that the full brightness alignment strategy causes a monotonic increase in LPIPS from 0.191 to 0.473, and PSNR decreases from 18.6 dB to 7.9 dB. These results demonstrate that the original loss term is crucial to the model's training process. Completely removing the original loss term severely impairs the model's performance.\\
\textbf{Does GT-Mean Loss achieve simultaneous improvement in perception and fidelity?} GT-Mean Loss achieves optimal LPIPS, PSNR, and mean brightness discrepancy. Its dynamic weighting mechanism $W$ adaptively adjusts the original loss term and brightness alignment term, avoiding the limitations of a single objective. Unlike the phase fluctuations of hybrid strategies or the collapse associated with full brightness alignment, GT-Mean Loss maintains continuous positive optimization of both perception and fidelity throughout training, leading to a transition from conflict to mutual enhancement.
\begin{figure}[htb]  
	\centering
	\includegraphics[width=0.65\linewidth]{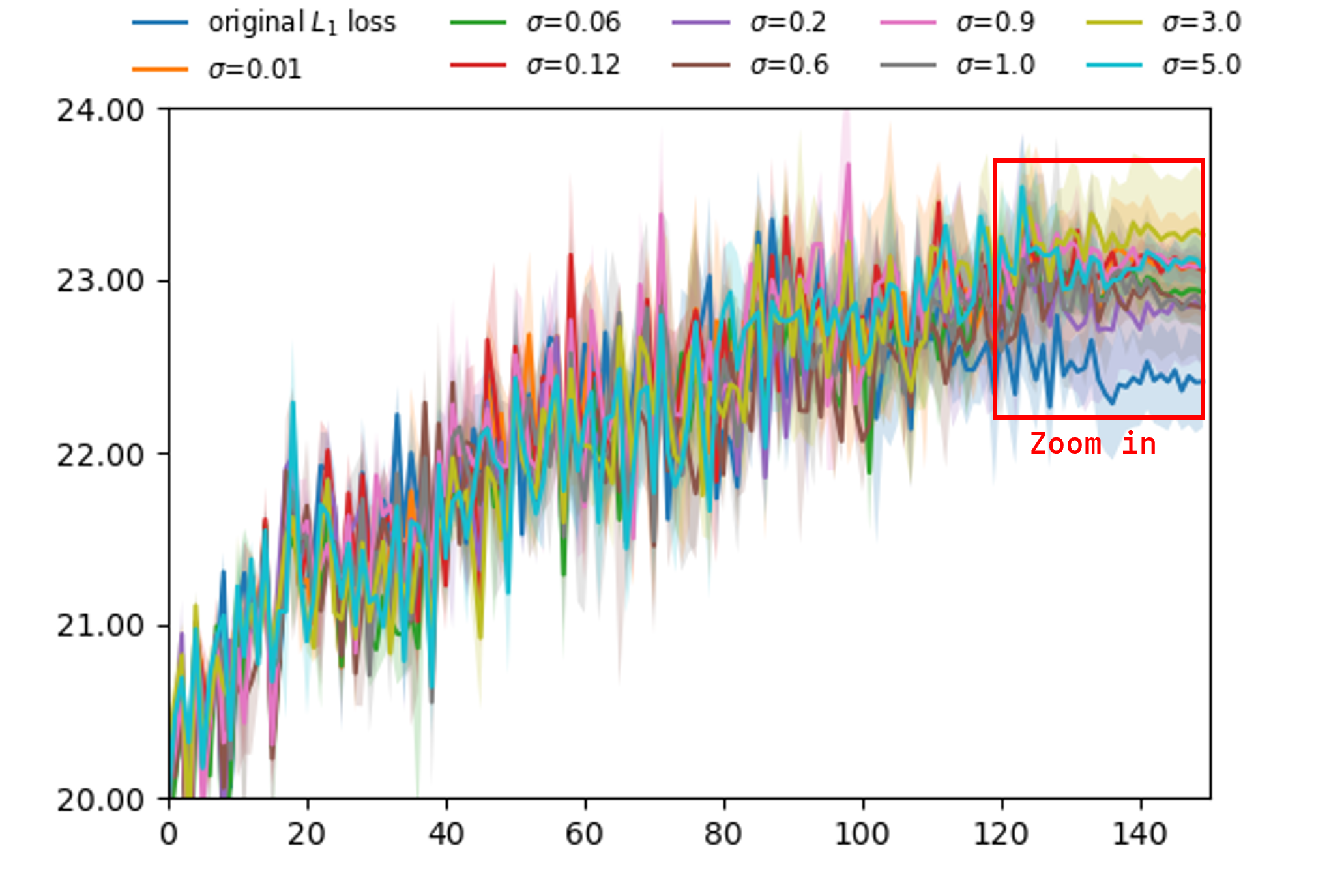}
		\centering
		\centering
		\text{(a) PSNR curves for 150K iterations}
		\centering
		\includegraphics[width=0.65\linewidth]{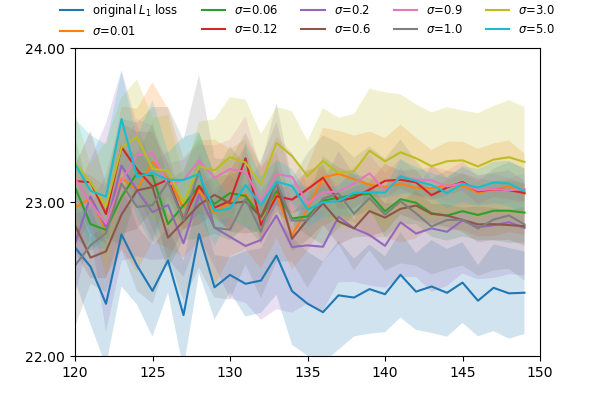}
		\centering
		\text{(b) PSNR curves last 30K iterations}
	\caption{The effect of different $\sigma$ on model performance.}
	\label{sigma_effect}
\end{figure}
\begin{figure*}[htb]
    \centering
    \begin{subfigure}[b]{0.25\textwidth} 
    \includegraphics[width=\textwidth]{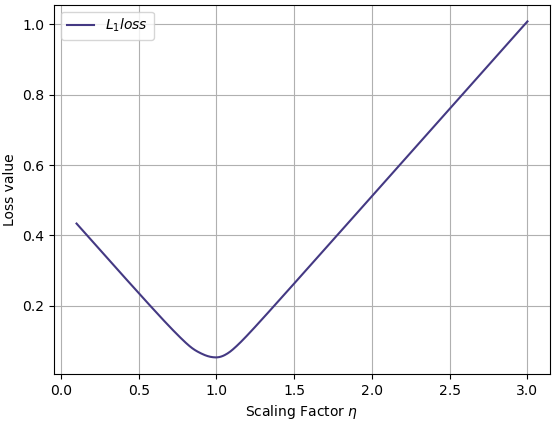}
        \caption{$L_1$ loss curve}
        \label{fig:loss_l1}
    \end{subfigure}
    \begin{subfigure}[b]{0.25\textwidth}\includegraphics[width=\textwidth]{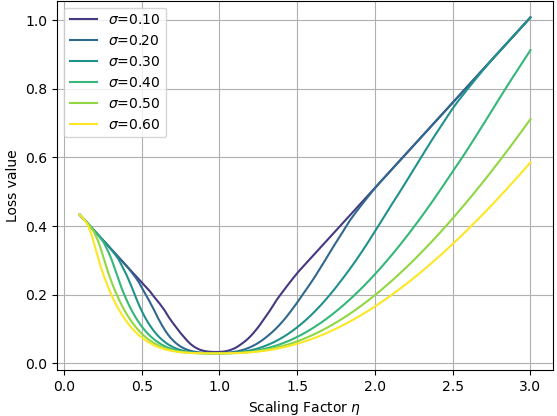}
        \caption{GT-mean $L_1$ loss curves}
        \label{fig:loss_gtmean}
    \end{subfigure}
    \begin{subfigure}[b]{0.25\textwidth} 
        \includegraphics[width=\textwidth]{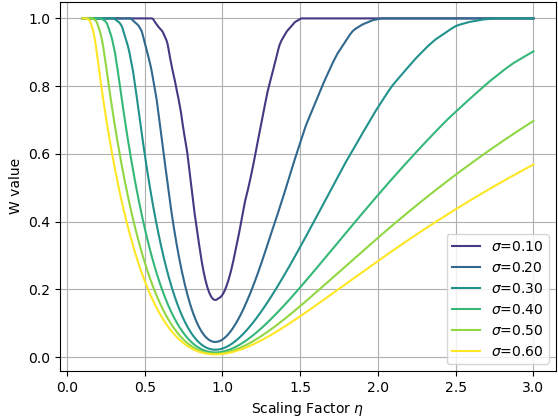}
        \caption{Curves of the weight $W$}
        \label{fig:weight_curve}
    \end{subfigure}
    \caption{Loss curves and weight curves for analyzing the effectiveness of the GT-mean loss.}
    \label{form}
\end{figure*}
\subsubsection{Qualitative Analysis.}
In Figure~\ref{fig4}B, we recorded the absolute difference $|y - f(x)|$ as residual heatmaps at 50K, 100K, and 150K iterations. The GT-Mean Loss demonstrates significantly smaller residuals in both texture and noise regions compared to the baseline, suggesting its superior capability in detail refinement.

In Figure~\ref{fig4}C, the final results generated using GT-Mean Loss exhibit excellent alignment with the ground truth (GT) in terms of color fidelity and structural clarity. In contrast, the baseline method shows noticeable color distortions, such as red artifacts, and fails to fully suppress noise.

\subsection{Effect of the parameter $\sigma$}
\label{Ablation Study}
To investigate the influence of $\sigma$, we conducted experiments on LOLv1 using RetinexFormer trained with GT-mean $L_1$ loss under different $\sigma$ values. We selected 10 different $\sigma$ values, running each setting three times for consistency. Notably, $\sigma=0$ represents a special case where the GT-mean $L_1$ loss degrades to the original $L_1$ loss. For every 1,000 (1K) iterations in the 150K iterations, we calculated mean and variance of PSNR, shown in Figure \ref{sigma_effect} for demonstrating the trend during training. In the early stages (as can be seen in Figure \ref{sigma_effect} (a)), the curve tendencies under different $\sigma$ settings are similar. Considering the curve with $\sigma=0$ is the original $L_1$ loss, we can verify that the GT-mean loss at the early stage behaves like the original $L_1$ loss. In contrast, as shown in the zoomed-in views of the last 30K iterations (Figure \ref{sigma_effect} (b)), we observe that all settings become stable, and the settings with non-zero $\sigma$ consistently perform better than $\sigma=0$. This observation shows that the GT-mean loss diverges significantly from the original $L_1$ loss in the late training stage. The second term in Eq.\ref{eq1} allows the GT-mean loss to continuously improve model performance. In addition, the experiment shows that the choice of $\sigma$ value is open. As $\sigma$ measures the spread of the random variable $\widetilde{\E}[\cdot]$ deviating from the observed average image brightness $\E[\cdot]$ in our modeling, we recommend using a small value, such as $\sigma=0.1$ for real world application. 

\subsection{Further analysis on the GT-mean loss}
\label{loss_form_chapter}
In this experiment, we further investigate the difference between the original loss and the GT-mean loss. Specifically, we randomly selected a batch of low-light images (batchsize = 8) and their corresponding ground truth images. These images were enhanced using RetinexFormer to produce enhanced outputs $f(x)$. To simulate the varying brightness, we introduced a unified scaling factor $\eta$ ranging from 0 to 3, simulating the progression of the enhanced images $\eta \cdot f(x)$ from dark to bright. This experiment setting simulates how the loss value varies under different brightness levels, facilitating us to investigate the loss curve with respect to the brightness variation only.

Based on the above experimental design, we present the curve of the original $L_1$ loss (Figure \ref{form}(a)), and the curves of the GT-mean $L_1$ loss under different $\sigma$ values (Figure \ref{form}(b)). The difference between them is that the use of the GT-mean loss clearly produces basins around $\eta = 1$. In another word, the GT-mean loss produces small-gradient region with regard to brightness around $\eta = 1$, of which the range is controlled by $\sigma$. From an optimization perspective, since the gradients with respect to brightness become smaller, the optimization along the direction of brightness adjustment is in turn slowed down. Based on this characteristic, in real-world model training, the GT-mean $L_1$ loss enables LLIE models to focus on other important degeneration factors, when $p(\widetilde{\E}[y])$ and $q(\widetilde{\E}[f(x)])]$ become closer. In contrast, the original $L_1$ loss is less capable of decoupling the optimization with respect to brightness and other visual quality factors.

Additionally, Figure \ref{form}(c) presents the weight curves that correspond to Figure \ref{form}(b), demonstrating how the GT-mean $L_1$ loss behaves with regard to weight variation. As $\eta$ approaches 1, the weight $W$ rapidly decreases, indicating that the second term in Eq.\ref{eq1} begins to dominate the loss function, confirming the mechanism of our loss. Notably, as $\sigma$ increases, $W$ starts to drop at smaller values of $\eta$, meaning that the second term in Eq.\ref{eq1} takes over earlier in the optimization process. This behavior aligns with the design of $\sigma$, which controls the spread of $\widetilde{\E}[\cdot]$.

\section{Conclusion}
In this paper, we propose the GT-mean loss to advance research on supervised low-light image enhancement (LLIE) methods. The GT-mean loss enables the model training process to circumvent the misleading issue caused by brightness mismatch, thereby comprehensively addressing the various degeneration factors in low-light images. Due to its simple construction, the GT-mean loss can be easily adopted by a wide range of  supervised LLIE methods, imposing minimal additional computational overhead during training. Experiments across various supervised LLIE methods consistently demonstrate the effectiveness of the proposed loss. 
While the estimation of the weight $W$ remains an open problem, we plan to explore various $W$-estimation strategies to potentially unlock even greater performance gains in LLIE models.

{
    \small
    \bibliographystyle{ieeenat_fullname}
    \bibliography{main}
}

\newpage
\ifappendix
\section*{Appendix}
\appendix
\section{EXPERIMENTAL DETAILS} 
\label{experiment_settings}

In this section, we present the experimental setup for each method. Our aim is to ensure consistency with the official settings for each baseline model while introducing the GT-mean loss to demonstrate its effectiveness. To ensure fair comparisons, both the baseline models and the ones using GT-mean loss were trained under identical hardware and software environments, minimizing the effects of randomness.

\textbf{Uformer.}  
Both the baseline and the GT-mean loss variant were trained following the experimental setup for motion deblurring in \citep{Wang_2022_CVPR}, selected Uformer-T as the backbone model. The Charbonnier loss used in the baseline was extended to GT-mean loss for the variant.

\textbf{MIRNet.}  
Both the baseline and the GT-mean loss variant were trained according to the settings used for the denoising task in \citep{Zamir2020MIRNet}. In the GT-mean loss variant, the Charbonnier loss was replaced with the GT-mean loss.

\textbf{RetinexFormer.}  
For both the baseline and the GT-mean loss variant, we followed the training settings for LOL datasets in \citep{Cai_2023_ICCV}. The $L_1$ loss used in the baseline was extended to GT-mean loss in the variant.

\textbf{Restormer.}  
The baseline and the GT-mean loss variant were both trained following the motion deblurring settings described in \citep{Zamir2021Restormer}. The $L_1$ loss in the baseline was extended to GT-mean loss in the variant.

\textbf{LLFormer.}  
Both the baseline and the GT-mean loss variant were trained according to the settings for the LOLv1 dataset described in \citep{wang2023ultra}. The Smooth $L_1$ loss used in the baseline was extended to GT-mean loss for the variant.

\textbf{SNR-Aware.}  
The baseline and the GT-mean loss variant were both trained using the settings for for LOL datasets outlined in \citep{9878461}. The Charbonnier loss and perceptual loss used in the baseline were extended to GT-mean loss in the variant.

\textbf{CID-Net.}  
Both the baseline and the GT-mean loss variant were trained using the LOLv1 settings described in \citep{feng2024hvi}. In the GT-mean loss variant, the Charbonnier loss, edge loss, and perceptual loss were extended to GT-mean loss.

In summary, for each method, the original loss functions were extended to GT-mean loss, and all models were trained using consistent settings to ensure a fair comparison.
\fi
\end{document}